\documentclass{article}

\usepackage{microtype}
\usepackage{graphicx}
\usepackage{subcaption}
\usepackage{booktabs} 

\usepackage{hyperref}

\usepackage[accepted]{icml2026}

\usepackage{amsmath}
\usepackage{amssymb}
\usepackage{mathtools}
\usepackage{amsthm}
\usepackage{multirow}
\usepackage{tabularx}
\usepackage{threeparttable}
\usepackage{xcolor}
\usepackage[table]{xcolor}
\usepackage{etoc}
\usepackage{makecell}

\usepackage[utf8]{inputenc}
\usepackage[T1]{fontenc}

\usepackage[capitalize,noabbrev]{cleveref}

\theoremstyle{plain}

\theoremstyle{definition}

\theoremstyle{remark}

\usepackage[textsize=tiny]{todonotes}

\usepackage{xspace}
\usepackage{colortbl}
\usepackage{longtable}
\usepackage{lineno}

\usepackage{fontawesome}

\newcommand{\ours}{DeepMedix\xspace}
\newcommand{\oursreasoning}{DeepMedix-R1\xspace}
\newcommand{\qwen}{Qwen2.5-VL-7B\xspace}
\newcommand{\qwenthink}{Qwen2.5-VL-7B (think)\xspace}
\newcommand{\qwenlarge}{Qwen2.5-VL-32B\xspace}
\newcommand{\deepseek}{DeepSeek-VL-7B\xspace}
\newcommand{\huatuo}{HuatuoGPT-Vision\xspace}
\newcommand{\chexagent}{CheXagent\xspace}
\newcommand{\llavarad}{LLaVA-Rad\xspace}
\newcommand{\medgemma}{MedGemma\xspace}
\newcommand{\blue}{\textcolor{blue}}

\newcommand{\snow}{\includegraphics[height=0.8em]{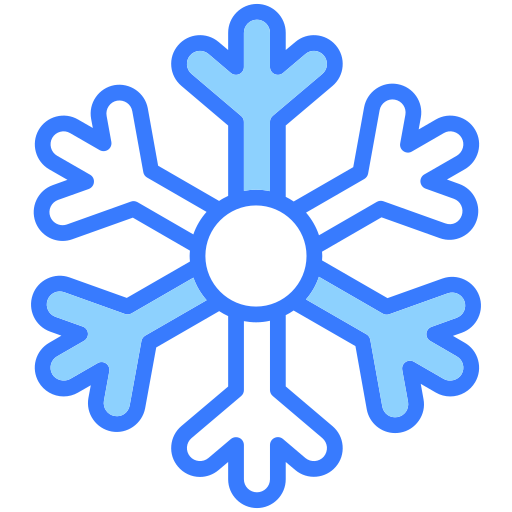}}
\newcommand{\fire}{\includegraphics[height=0.8em]{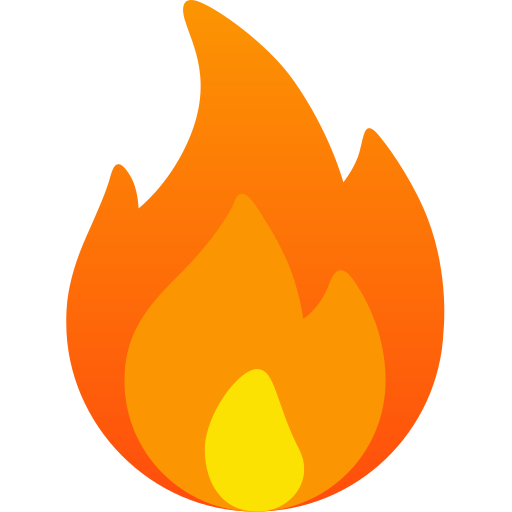}}

\newcommand{\mimicfindings}{MIMIC-CXR findings\xspace}
\newcommand{\mimicimpression}{MIMIC-CXR impression\xspace}
\newcommand{\openifindings}{OPEN-I findings\xspace}
\newcommand{\openiimporession}{OPEN-I impression\xspace}
\newcommand{\extvqa}{Ext-VQA\xspace}
\newcommand{\cxrvqa}{CXR-VQA\xspace}
\newcommand{\cxrnew}{CXR14\xspace}

\newcommand{\myitem}{\noindent\textbf{\emph{$\bullet$}}}

\icmltitlerunning{Toward Clinically Explainable AI for Medical Diagnosis}

\begin{document}

\twocolumn[
  \icmltitle{Toward Clinically Explainable AI for Medical Diagnosis: A Foundation Model with Human-Compatible Reasoning via Reinforcement Learning}
    
    \vspace{-0.7em}
    \centerline{
        \textbf{Qika Lin}$^{1}$ \quad 
        \textbf{Yifan Zhu}$^{2}$ \quad 
        \textbf{Bin Pu}$^{3}$ \quad 
        \textbf{Ling Huang}$^{4}$ \quad 
        \textbf{Haoran Luo}$^{5}$ \quad 
        \textbf{Jingying Ma}$^{1}$ \quad 
        \textbf{Feng Wu}$^{1}$
    }
    \centerline{
        \textbf{Kai He}$^{1}$ \quad 
        \textbf{Jiaxing Xu}$^{1}$ \quad 
        \textbf{Zhen Peng}$^{6}$ \quad 
        \textbf{Tianzhe Zhao}$^{6}$ \quad 
        \textbf{Fangzhi Xu}$^{6}$ \quad
        \textbf{Jian Zhang}$^{6}$ 
    }
    \centerline{
        \textbf{Zhonghong Ou}$^{2}$ \quad 
        \textbf{Erik Cambria}$^{5}$ \quad 
        \textbf{Swapnil Mishra}$^{1}$ \quad 
        \textbf{Mengling Feng}$^{1}$
    }
    \centerline{$^{1}$National University of Singapore \quad $^{2}$Beijing University of Posts and Telecommunications}
    \centerline{$^{3}$The Hong Kong University of Science and Technology \quad $^{4}$Imperial College London}
    \centerline{$^{5}$Nanyang Technological University \quad $^{6}$Xi'an Jiaotong University}
    \centerline{\texttt{qikalin@foxmail.com, ephfm@nus.edu.sg}}

  \vskip 0.15in
]

\makeatletter
\global\icml@noticeprintedtrue
\makeatother




\begin{abstract}
The clinical adoption of artificial intelligence (AI) in medical diagnostics is critically hampered by its \emph{black-box} nature, which prevents clinicians from verifying the rationale behind automated decisions. To overcome this fundamental barrier, we introduce \oursreasoning, a foundation model (FM) for chest X-ray (CXR) interpretation that generates not only accurate diagnoses but also a transparent, step-by-step reasoning process grounded in specific visual evidence.
Our methodology employs a sequential training strategy, beginning with instruction fine-tuning, followed by a cold-start phase to elicit reasoning capabilities. Critically, we then implement reinforcement learning with grounded rewards to meticulously refine the model, aligning both its diagnostic outputs and its reasoning pathways with clinical plausibility.
Quantitative assessments show that \oursreasoning substantially outperforms advanced FMs, achieving improvements in report generation and visual question answering tasks.
We also introduce Report Arena, a novel LLM-based benchmark that ranks DeepMedix-R1 first among competing models for output quality.
Most significantly, a formal review by clinical experts reveals a profound preference for \oursreasoning's generated reasoning over the broadly adopted Qwen2.5-VL-7B model, confirming its superior interpretability and clinical utility.

\end{abstract}

\section{Introduction}

\begin{figure}[t!]
    \centering
        \includegraphics[width=0.9\linewidth]{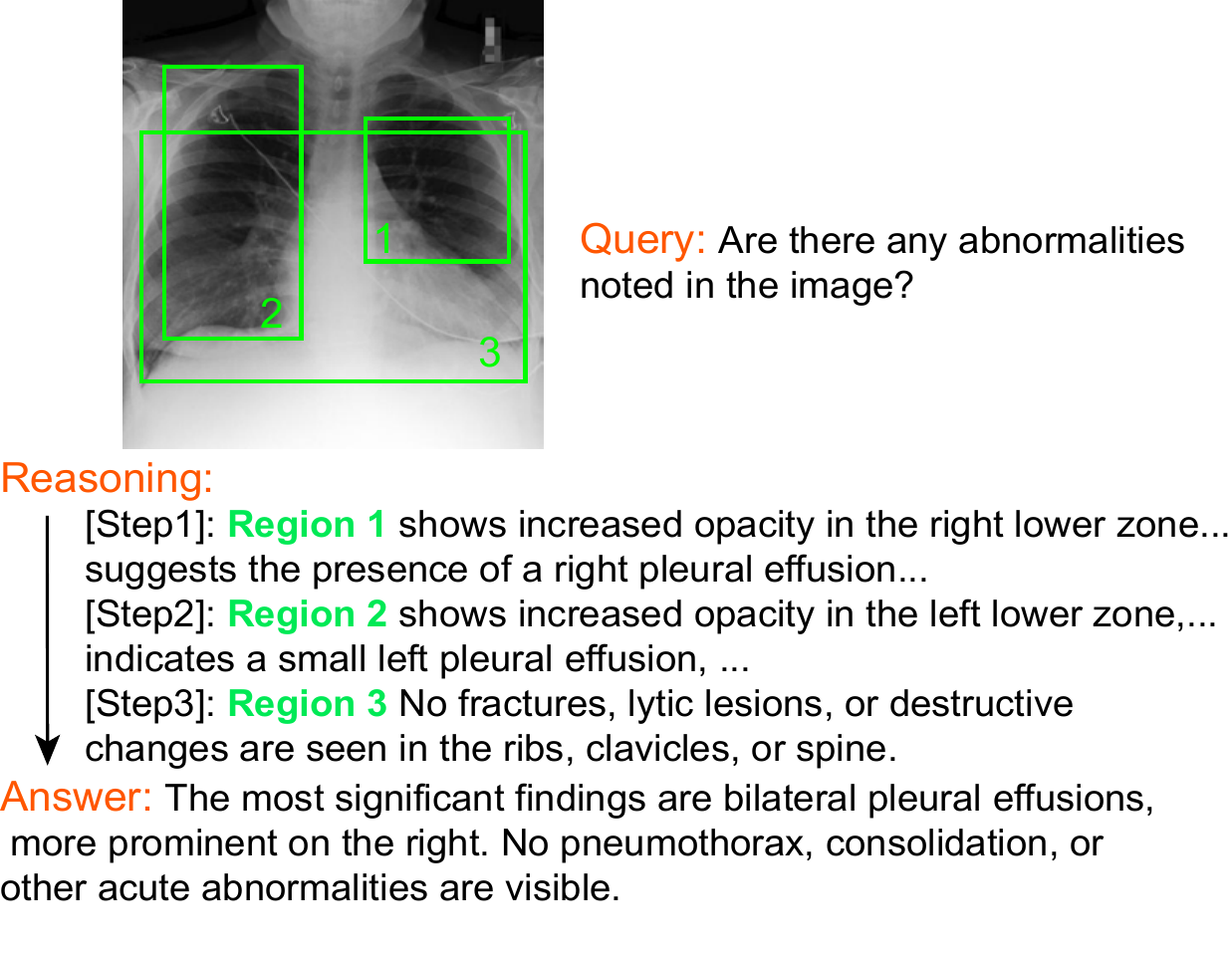}
        \vspace{-0.75em}
    \caption{The illustration of a step-by-step reasoning process grounded in specific visual regions for diagnostic decision-making.
    }
    \label{fig_intro1}
    \vspace{-0.75em}
\end{figure}

The rapid evolution of generative artificial intelligence (AI), particularly foundation models (FMs)—including both large language models (LLMs)~\cite{zhao2023survey} and vision-language models (VLMs)~\cite{yin2024survey}, promises to reshape the landscape of modern medicine~\cite{moor2023foundation,jiiret,lin2024has,neidlinger2025benchmarking,yang2025demographic}.
These models, trained on vast, multimodal datasets, have demonstrated remarkable capabilities in diverse healthcare applications, from interpreting medical images~\cite{chen2024towards} to analyzing electronic health records~\cite{guo2024multi}. By leveraging a single, adaptable architecture, FMs offer a powerful paradigm for enhancing diagnostic accuracy, streamlining clinical workflows, and accelerating medical discovery~\cite{thirunavukarasu2023large,zhang2024generalist,singhal2025toward}.
Recent studies have produced specialized models for medicine and healthcare applications, such as LLaVA-Med~\cite{li2023llava} for answering research questions, ALPaCA~\cite{gao2025alpaca} for computational pathology,
and LLaVA-Rad~\cite{zambrano2025clinically} for chest X-ray (CXR) report generation, underscoring the field's rapid progress.

\begin{figure*}[t!]
    \centering
        \includegraphics[width=0.96\linewidth]{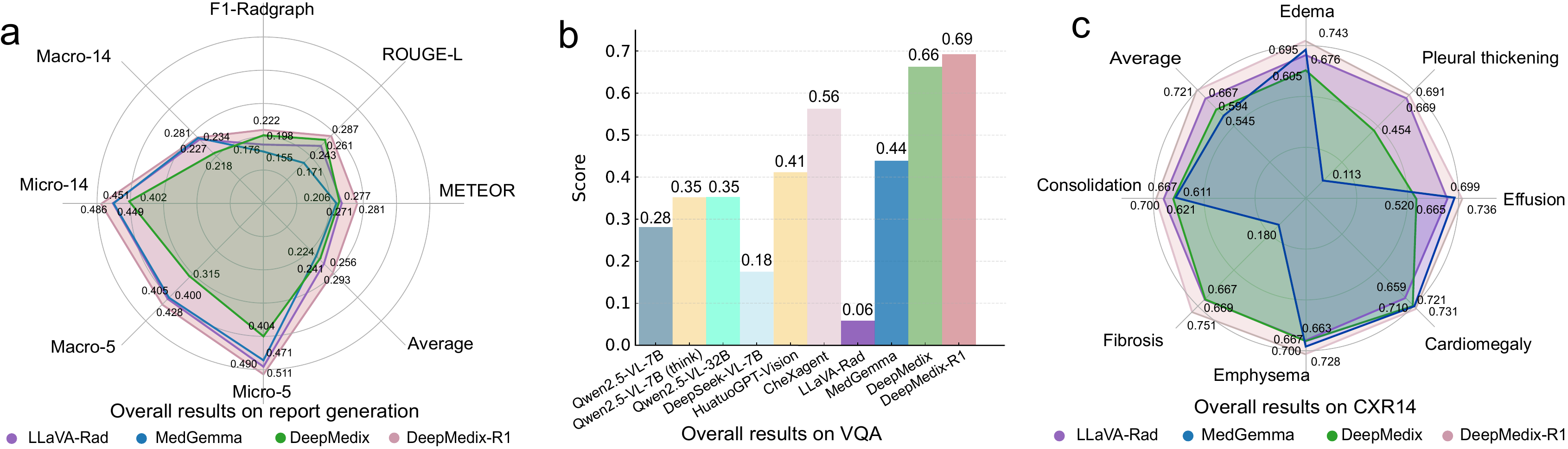}
    \caption{The overall experimental results. (a), (b) and (c) are the results for model comparison on report generation, VQA, and external CXR14 dataset, respectively, demonstrating \oursreasoning outperforms strong open-source FMs in both general and medical domains.
    }
    \label{fig_intro2}
    \vspace{-0.75em}
\end{figure*}

However, a formidable obstacle impedes the translation of these models into routine clinical practice: the absence of human-compatible transparent reasoning. Current medical FMs predominantly function as \emph{\textbf{black boxes}}, delivering directly outputs without exposing the underlying clinical logic~\cite{mesinovic2025explainability}. This lack of interpretability is a critical flaw in high-stakes medical environments, where the ability for a clinician to scrutinize and validate an AI's decision-making process is paramount for patient safety and professional accountability~\cite{lievin2024can,lai2025med}.
This challenge is compounded by a second limitation, i.e., the frequent inability of models to precisely ground their findings in localized regions of an image~\cite{bannur2024maira,santiesteban2024enhancing,chen2025large}. Since radiological diagnosis relies fundamentally on the meticulous evaluation of specific anatomical and pathological features, this lack of granular, visually-grounded analysis represents a critical capability gap~\cite{dong2025generative,de2025padchest}.
For instance, a model may correctly predict the presence of \emph{pneumonia} but fail to associate its decision with the corresponding opacity in the lower lobe of the lung, thereby offering limited diagnostic reliability in clinical application and overlooking the critical regions that should guide subsequent clinical attention.

To address these limitations, we propose a novel approach integrating human-compatible grounded interpretation and step-wise reasoning for medical diagnosis using FMs to mimic the expert decision-making process, with a specific focus on CXRs, as shown in Fig.~\ref{fig_intro1}.
First, we systematically curate a comprehensive multi-source CXR dataset encompassing diverse clinical scenarios, including report generation (\emph{findings} and \emph{impression} sections), visual question answering (VQA), and localized image interpretation. Building upon this foundation, we adapt the advanced general-purpose VLM \qwen~\cite{qwen2.5-VL} through extensive instruction fine-tuning (351.5K samples), equipping the resulting \ours model with fundamental CXR interpretation capabilities.
Second, we generate high-quality synthetic reasoning data (3.9K samples) incorporating grounded image representations to overcome the model cold-start problems~\cite{shao2024deepseekmath,wei2025advancing}.
The continuous fine-tuning upon it endows the model with preliminary capabilities in grounded understanding and reasoning.
Finally, we implement an advanced online reinforcement learning (RL) framework, i.e., group relative policy optimization (GRPO)~\cite{shao2024deepseekmath}, to simultaneously enhance both reasoning quality and generation performance of \ours via proposed grounded rewards,
ultimately creating \oursreasoning model.

To comprehensively evaluate the CXR interpretation capabilities of open-source FMs across both general and medical-specific domains, we develop \textbf{XrayBench} (as shown in Fig.~\ref{fig_bench}, including internal and external data), a multimodal evaluation framework comprising 16.3K carefully curated samples. It systematically assesses model performance on two internal clinical tasks: report generation (findings and impression sections), and VQA.
To assess the quality of the final output, conventional natural language generation (NLG)-based metrics and fact-based evaluation are utilized.
Besides, an LLM-as-judge evaluation framework for comprehensive performance analysis, namely \textbf{Report Arena}, is introduced to assess paired outputs from competing models on report generation tasks.
The CXR14~\cite{wang2017chestx} dataset is employed as external diagnostic data.
The overall results are shown in Fig~\ref{fig_intro2}.
Additionally, CXR experts conduct a comprehensive evaluation of the reasoning process.
We have three key observations from the experiment results:

\myitem\;On automated metrics for both report generation
and VQA,
\oursreasoning outperforms state-of-the-art open-source FMs in both general and medical domains, achieving 31.32\% and 57.75\% higher than MedGemma~\cite{medgemma-hf} on report generation and VQA tasks, respectively.
Besides, \oursreasoning demonstrates relative improvements of 21.63\% over CheXagent and 8.24\% over LLaVA-Rad when evaluated on the external diagnostic dataset.

\myitem\;We implement an LLM-as-judge evaluation framework (i.e., Report Arena) to comparatively assess paired outputs from competing models on report generation tasks.
Benchmarking results position \oursreasoning in first place, followed sequentially by LLaVA-Rad~\cite{zambrano2025clinically}, \ours, and MedGemma models.

\myitem\;We carry out annotation by four domain experts who evaluate the reasoning processing across five key dimensions, i.e., \emph{Relevance}, \emph{Correctness}, \emph{Completeness}, \emph{Grounded Preference}, and \emph{Overall Preference}. Evaluation results demonstrate that \oursreasoning generally outperforms Qwen2.5-VL-7B in reasoning quality and is strongly preferred by medical experts.


\section{Methodology}

\subsection{Model Optimization}

Like LLM fine-tuning, in the first stage for instruction fine-tuning and the second stage for cold start reasoning, the next token prediction~\cite{chen2024next} is carried out to optimize FMs based on the instruction $\mathcal{Q}$ and X-ray $\mathcal{I}$ and previously generated text tokens with the loss function: 
\begin{equation}
  \mathcal{L}_{\rm SFT}=-\sum_{n=1}^N \log\big(x_n|x_{<n},\mathcal{I},\mathcal{O}\big).
\end{equation}
In the second stage, we prompt closed-scoured GPT-4.1~\cite{openai2023gpt4} to generate reasoning processing given X-rays, task prompt, and ground truth. The detailed design of the prompt is shown in Fig.~\ref{fig_prompt}. We then check the validity of the derived reasoning procedures by prompting Qwen2.5-VL-32B and filter out those that are clearly unreasonable, e.g., the provided bounding box of the image region exceeds the maximum edge of the image. Finally, 3.9K samples are kept and used for the cold start training.

In the third stage, the GRPO algorithm~\cite{shao2024deepseekmath} is utilized for online RL to enhance the reasoning ability.
Specifically, we optimize the model by comparing the relative performance of different policies or actions within the same group for CXR tasks,
with a specific focus on the perception of grounded visual regions.
It is to enable the model to compare its output with multiple candidate outputs generated by itself.
Formally, $P(Q)$ denote the question set used for training.
$q=[\mathcal{Q},\mathcal{I}]$ is a sampled query in the current iteration.
$\pi_{\theta_{\rm old}}$ and $\pi_{\theta_{\rm new}}$ ($\pi_\theta$) denote the old policy and current new policy, respectively.
$o$ is a complete response sampled from a policy. Let $\pi_{\theta_{\rm ref}}$ denote the reference policy, which in practice is the frozen base model (DeepMedix).
$G$ is the number of responses sampled per question in each iteration. 
For each question $q$, GRPO samples a group of outputs $\{o_1,o_2,\cdots,o_G\}$ from the old policy model $\pi_{\theta_{\rm old}}$. The policy model $\pi_\theta$ is optimized by maximizing the objective:
\begin{equation}
\small
\begin{split}
    &\mathcal{J}_{GRPO}(\theta) = \mathbb{E}{[q \sim P(Q), \{o_i\}_{i=1}^G \sim \pi_{\theta_{old}}(O|q)]}  \\
    & \frac{1}{G}\sum_{i=1}^G\frac{1}{|o_i|} \sum_{t=1}^{|o_i|} \left\{ \min \left[ r_{i,t}(\theta) \hat{A}_{i,t}, \text{clip} \left( r_{i,t}(\theta), 1 - \epsilon, 1 + \epsilon \right)  \hat{A}_{i,t} \right] \right. \\
    & \left. - \beta \mathbb{D}_{KL}\left[\pi_{\theta} || \pi_{ref}\right]\right\}, \text{where}\;r_{i,t}(\theta)=\frac{\pi_\theta(o_{i,t} | q, o_{i,<t})}{\pi_{\theta_{old}}(o_{i,t} | q, o_{i,<t})}.
\end{split}
\label{eq:GRPO-obj}
\end{equation}
$\epsilon$ is a hyper-parameter.
The KL divergence~\cite{kullback1951information} term serves as a regularization for the policy update, preventing $\pi_\theta$ from deviating too far from the reference model $\pi_{\theta_{old}}$.
$\beta$ is the coefficient of the KL penalty.
$\hat{A}_{i,t}$ represents the advantage computed using relative rewards among outputs within each group.
Suppose a group of outputs $\{o_1,o_2,\cdots,o_G\}$ has reward $\mathbf{r}=\{r_1,r_2,\cdots,r_G\}$, there is $\hat{A}_{i,t}=\frac{r_i-{\rm mean}(\mathbf{r})}{{\rm std}(\mathbf{r})}$.
For each output $o_i$ and its ground label $l_i$, we set its reward $r_i$ based on three distinct perspectives, i.e., answer score, image grounded score, and format score. The answer score directly reflects the alignment degree between the generated answer and the ground truth, given by:
\begin{equation}
\small
    r_{a,i}=1.5*
\left
\{
\hspace{-5pt}
\begin{array}{lcl}
\mathbb{I}(o_i=l_i),\quad\text{closed-ended questions} \\
{\rm F}1(o_i, l_i),\quad\text{multi-object questions} \\
1/3[{\rm B1}(o_i, l_i) + {\rm B4}(o_i, l_i) + {\rm RL}(o_i, l_i)],\text{else}. \\
\end{array} 
\right.
\end{equation}
B1, B4, and RL are short for BLEU-1, BLEU-4~\cite{DBLP:conf/acl/PapineniRWZ02}, and ROUGE-L~\cite{lin2004rouge}, respectively.
Image grounded scores focus the model's awareness on specific local grounded regions within CXRs, given by:
\begin{equation}
    r_{g,i}=\max \big({\rm N}_{coo}(o_i)*0.05+\phi(o_i), 0.15\big).
\end{equation}
${\rm N}_{coo}$ function counts the number of coordinates presented in the reasoning process. $\phi$ is the deviation control parameter, and we set it to -0.2, which means that we set a penalty if the coordinate is beyond the range of the CXR image.
The format score indicates if the model response followed the predefined format, i.e., the thinking process in the ``<think>...</think>'' tag and the final result in ``\textbackslash  boxed\{\}'' tag, given by:
\begin{equation}
    r_{f,i}=
\left
\{
\begin{array}{rcl}
1.0 && \text{conforms to format} \\
0.0 && \text{else}. \\
\end{array} 
\right.
\end{equation}
Finallly, the final reward is formally as: $r_i=(1-\lambda)r_{a,i}+\lambda r_{f,i}+r_{g,i}$. The $\lambda$ is set to 0.1 as the default.

\subsection{Report Arena}

To give a comprehensive evaluation of cutting-edge open-sourced vision-language FMs, automated NLG metrics (BLEU, METEOR~\cite{DBLP:conf/wmt/LavieA07}, and ROUGE-L) and fact metrics (F1-RadGraph~\cite{DBLP:conf/nips/JainASTDBC0LNLR21} and F1-CheXbert~\cite{smit2020combining}) are utilized on report generation tasks.
In the VQA tasks, we utilize the overall accuracy covering both exact match accuracy of closed-ended questions and F1 scores for \emph{multi-object} questions that need to predict a set of abnormalities, positions or attributes.
Beyond these automated metrics for evaluation, inspired by Chatbot Arena~\cite{chiang2024chatbot,cheng2025caparena}, we implement an LLM-as-judge evaluation framework, namely Report Arena to comparatively assess paired outputs from competing models on report generation tasks. 
Suppose there are totally M models, $m_1,m_2\in[{\rm M}]$ are the indices of the models.
We define $A_t=(m_1,m_2)$ as the model pair at time $t$, and the response from LLM is $H_t\in \{0,1\}$, indicating whether model $m_1$ is better than $m_2$ for a specific report generation given a CXR image.
In the current stage $t$, the probability $P_t(a)$ for sampling model pair $a$ for evaluation is given by:
\begin{equation}
\small
    P_t(a) \propto \sqrt{\frac{\hat{\sum}_{t,a,a}}{\left| \{ t : A_t = a \} \right|}} - \sqrt{\frac{\hat{\sum}_{t,a,a}}{\left| \{ t : A_t = a \} \right| + 1}},
\end{equation}
where $\hat{\sum}_{t}$ is the estimated covariance matrix.
After the sampling and judging process, the Bradley-Terry~\cite{bradley1952rank} model with the logistic form is implemented to calculate scores for all models:
\begin{equation}
\small
\hat{s} = \arg\min_{\xi} \sum_{t=1}^{T} \frac{1}{P(A_t)} {\rm CE}\big( H_t, \frac{1}{1 + e^{\xi_{A_{t,2}} - \xi_{A_{t,1}}}} \big),
\end{equation}
where $\xi$ is a {\rm M}-dimensional vector to represent model scores and {\rm CE} is the binary cross-entropy loss.
We normalize the learned ranking scores to the [0,1] range using max-absolute scaling.
Totally, 6K pairs are used for evaluation.

\subsection{Expert Annotation}

For the final reasoning processing annotation, we design five metrics covering three progressive levels:

(1) Reasoning step. Reflecting the detail of each reasoning step of reasoning processing, metric \emph{Relevance} and \emph{Correctness} are utilized. The former indicates whether the content described in the specific step is relevant to the correct answer, while the latter denotes whether the content described in the specific step truly reflects the CXR content, including specific local regions.

(2) The whole reasoning process of models. Metric \emph{Completeness} is utilized, indicating whether the reasoning process is complete, i.e., for obtaining the correct answer, are all the reasoning process descriptions provided by the model sufficient?

(3) Reasoning preference for samples. Given the sample and the generated reasoning process by models, we need to judge which model is better. Metric \emph{Grounded Preference} and \emph{Overall Preference} are introduced. The former evaluates the fine-grained image description of the overall reasoning process and determines which model is better, covering whether there are fine-grained descriptions, their correctness, and their correlation regarding the correct answer.
The latter evaluates the professionalism of the overall reasoning process and the extent to which the correct answer can be obtained through reasoning.

For the annotation, we randomly sampled 210 samples and their reasoning processing generated by \oursreasoning and Qwen2.5-VL-7B (think) from XrayBench.
Four expert CXR physicians were divided into two groups for annotation, with each group responsible for evaluating all samples.
The detailed template for the annotation is shown in Fig.~\ref{fig_annotation}.

\section{Experiment Results}

\subsection{Overview of Results}
In the experiment, we employ three general-purpose vision-language FMs: \qwen~\cite{qwen2.5-VL}, \qwenlarge~\cite{qwen2.5-VL}, \deepseek~\cite{lu2024deepseek}.
To investigate the effects of grounded understanding and reasoning, we introduce \qwenthink that employs specialized prompting to enable the \qwen model to generate both the reasoning process and the final answer simultaneously.
For comparative analysis, we additionally evaluate two medical FMs (HuatuoGPT-Vision~\cite{chen2024huatuogptvisioninjectingmedicalvisual} and MedGemma~\cite{medgemma-hf}) and two CXR-specific FMs (CheXagent~\cite{chen2024chexagent} and LLaVA-Rad~\cite{zambrano2025clinically}).
The XrayBench for testing integrates five key datasets: two for report generation (MIMIC-CXR~\cite{johnson2019mimic} and Open-I~\cite{demner2012design}), two for VQA (\extvqa~\cite{bae2024mimic} and \cxrvqa~\cite{Hu2025cxrvqa}), and CXR14 for external evaluation.
Its detailed statistics are shown in Fig.~\ref{fig_bench}.

The overall results for report generation, VQA, and external diagnosis of \ours-R1 and baselines are shown in Fig.~\ref{fig_intro2} (a), (b), and (c), respectively.
For report generation, \oursreasoning outperforms all compared models, including the supervised fine-tuning baseline \ours, as well as domain-specific models LLaVA-Rad and MedGemma, across all evaluation metrics, with average improvements of 21.76\%, 14.54\%, and 31.32\%, respectively.
For VQA, \oursreasoning also performs best across all baselines, while LLaVA-Rad shows a relatively limited performance as it is specifically fine-tuned only for report generation tasks.
\oursreasoning also demonstrates the superiority on the external CXR14 dataset.

\begin{figure*}[t!]
    \centering
    \includegraphics[trim=150pt 0pt 150pt 0pt, clip, width=0.9\textwidth]{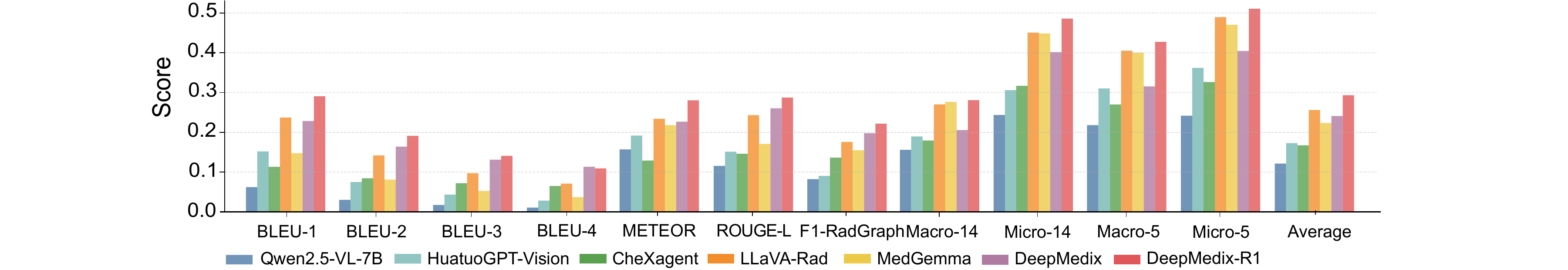}
        \vspace{-0.35em}
    \caption{
    Overall results for report generation tasks, representing the weighted average performance across all datasets.
    }
    \label{fig_rg1}
    \vspace{-0.75em}
\end{figure*}

\begin{figure*}[t!]
    \centering
        \includegraphics[width=0.9\linewidth]{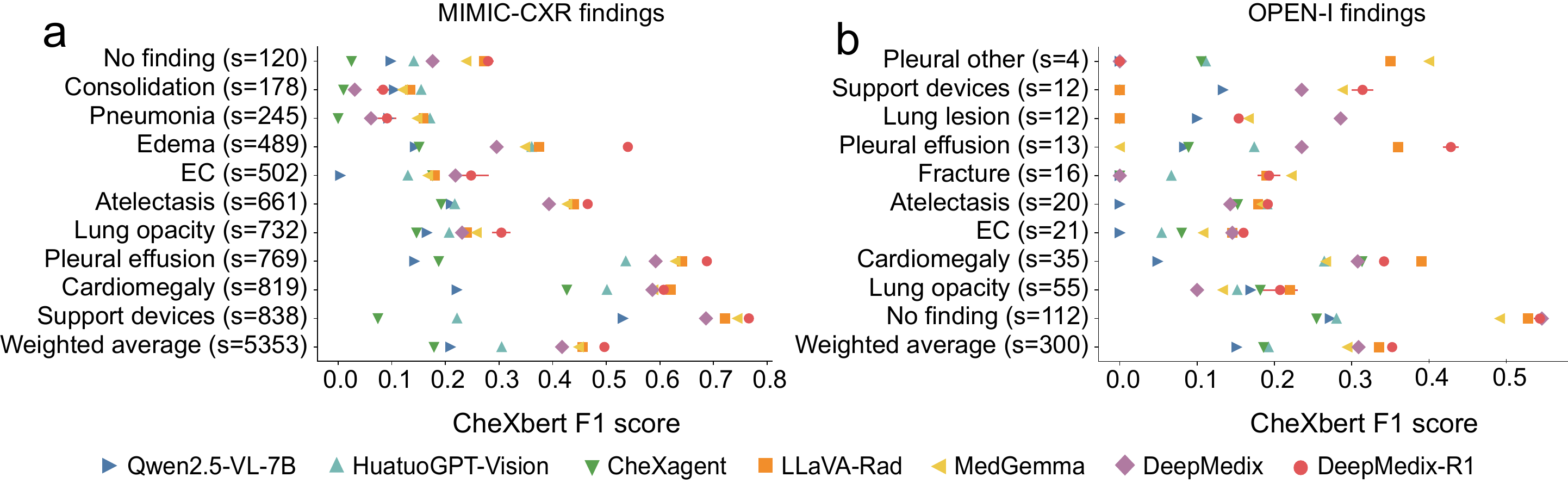}
        \vspace{-0.35em}
    \caption{
    The detailed CheXbert-F1 scores on top-10 supported observations on \mimicfindings and \openifindings, respectively. ``EC'' is short for ``Enlarged Cardiomediastinum''.
    }
    \label{fig_rg2}
    \vspace{-0.75em}
\end{figure*}

\subsection{Automated metrics on report generation}

The detailed results of the report generation are summarized in Tab.~\ref{tab_rg} and the average performance across four tasks is shown in Fig.~\ref{fig_rg1}.
For the detailed components of the test sets, i.e., \mimicfindings, \mimicimpression, \openifindings, and \openiimporession, it is evident that general-purpose models perform poorly across both NLG-based and fact-based evaluation metrics.
Among FMs in the medical domain, \llavarad serves as a strong baseline, particularly on the \mimicfindings and \mimicimpression sections. However, our proposed \oursreasoning achieves an average improvement of 10.20\% (0.3230 vs. 0.2931) and 10.47\% (0.2563 vs. 0.2320), respectively.
On average, we can observe that \oursreasoning consistently outperforms all other models, achieving the highest scores in most metrics and the overall highest weighted average of 0.2934.
Specifically, the general-purpose FMs (\qwen, \qwenlarge, and \deepseek) exhibit limited performance. The best model \qwenlarge among them has an average of 0.1433, which is only 28.84\% of \oursreasoning.
Also, the thinking process does not enhance the model's performance.
Compared with medical FMs, \oursreasoning has obvious improvements, i.e., 69.63\% for \huatuo (0.1730), 75.30\% for \chexagent (0.1674), 14.54\% for \llavarad (0.2562), 31.20\% for \medgemma (0.2236).
Furthermore, compared with fine-tuned \ours, \oursreasoning demonstrates consistent improvements across all four test components as well as the average part (0.2934 vs. 0.2410), underscoring the efficacy of RL in CXR report generation.

\begin{figure*}[t!]
    \centering
        \includegraphics[width=0.9\linewidth]{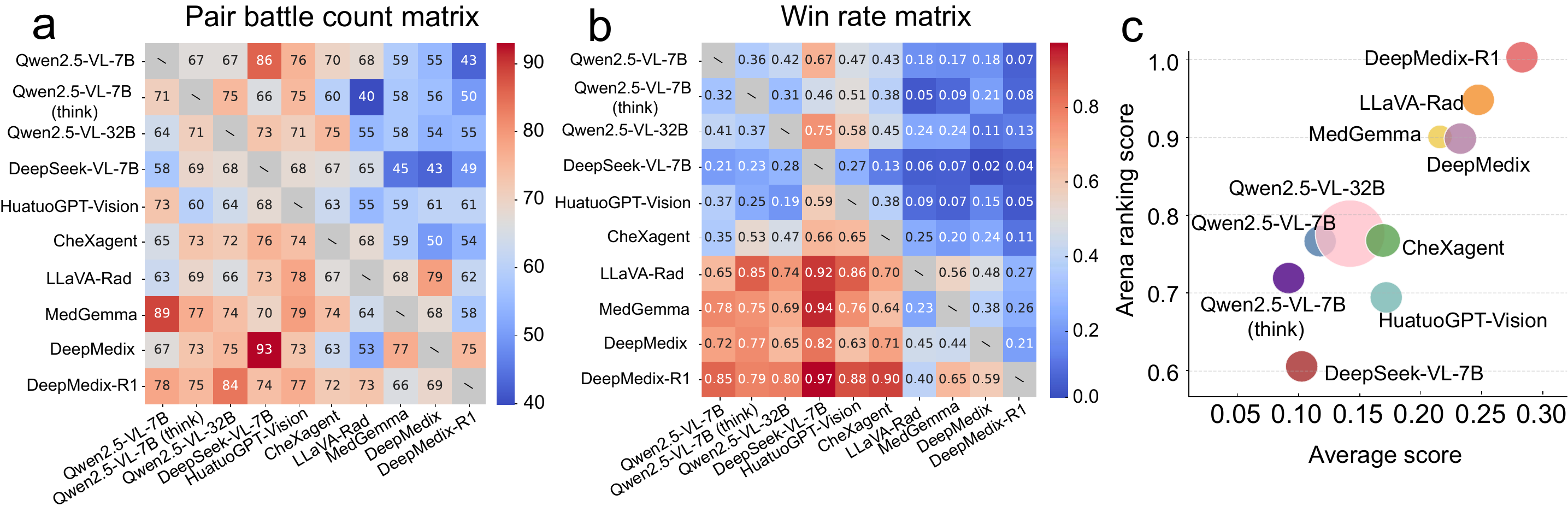}
        \vspace{-0.75em}
    \caption{
    The results of Report Arena.
    (a) and (b) present the numerical matrices for pairwise model comparisons and corresponding win rates in our Report Arena evaluation framework, respectively. (c) comparatively visualizes each model's performance across both conventional automated metrics and Report Arena rankings, with circle sizes proportionally representing the number of model parameters.
    }
    \label{fig_arena}
    \vspace{-0.75em}
\end{figure*}

For detailed evaluation, we provide the CheXbert F1 score for top-10 supported medical observations in the report collection, as shown in Fig.~\ref{fig_rg2} (a) and (b) on \mimicfindings and \openifindings, respectively.
\oursreasoning shows better F1 scores against baselines.
On average across \mimicfindings, \oursreasoning achieves a score of 0.4966, demonstrating significant improvements over existing FMs.
Specifically, it outperforms \qwen (0.2094) by 137.15\%, \huatuo (0.3050) by 62.80\%, \chexagent (0.1788) by 177.75\%, \llavarad (0.4562) by 8.86\%, \medgemma (0.4487) by 10.69\%, and even our baseline model (\ours, 0.4176) by 18.91\%.
Across all observations, \oursreasoning demonstrates optimal performance in detecting ``Support Devices'', ``Pleural Effusion'', ``Lung Opacity'', ``Atelectasis'', ``Enlarged Cardiomediastinum'', ``Edema'', and ``No Finding'' observations.
However, its capability is relatively limited for ``Cardiomegaly'', ``Pneumonia'', and ``Consolidation''.
On average across \openifindings, \oursreasoning achieves an F1 score of 0.3523, showing performance gains of 133.23\% over \qwen (0.1511), 83.10\% over \huatuo (0.1924), 89.52\% over \chexagent (0.1859), 5.03\% over \llavarad (0.3354), 19.71\% over \medgemma (0.2943), 14.09\% over \ours (0.3088).
Notably, while \ours initially fails to outperform \chexagent and \medgemma models, the implementation of online RL in \oursreasoning yields substantial performance gains, ultimately surpassing all benchmark models.

\subsection{Results of Report Arena}

Fig.~\ref{fig_arena} presents the pair battle count matrix, win rate matrix, and ranking scores from the report arena, respectively.
From Fig.~\ref{fig_arena} (b), models specifically designed for medical or CXR tasks (\llavarad, \medgemma, \ours, and \oursreasoning) consistently outperform general-purpose vision-language models (e.g., \qwen, \qwenlarge, and \deepseek) across most comparisons. For example, \ours achieves the 0.77 win rates against \qwenthink and 0.82 against \deepseek.
\huatuo does not perform well compared with other baselines, failing to outperform even general-purpose models and only having a marginal advantage over \deepseek with a win rate of 0.59.
\ours can win most baselines, except for \llavarad (win rate 0.45) and \medgemma (win rate 0.44).
Further, \oursreasoning can almost win all models, e.g., \huatuo (0.88), \chexagent (0.90), \medgemma (0.65).
While the (\oursreasoning, \llavarad) pair achieves a win rate of 0.4, the reverse pairing (\llavarad, \oursreasoning) shows a lower win rate of 0.27.
From Fig.~\ref{fig_arena} (c), it is observed that \oursreasoning ranks first, followed by \llavarad, \ours, \medgemma, \qwenlarge, \chexagent, \qwen, \qwenthink, \huatuo, and \deepseek in sequential order.
The results demonstrate that larger model size improves performance, while the reasoning mode of general-purpose models fails to yield significant gains.
Furthermore, we observe a moderately positive correlation between automated evaluation scores and ranking scores, with \oursreasoning achieving top performance across both metric types.

\subsection{Evaluation on VQA}

\begin{figure*}[t!]
    \centering
        \includegraphics[width=0.9\linewidth]{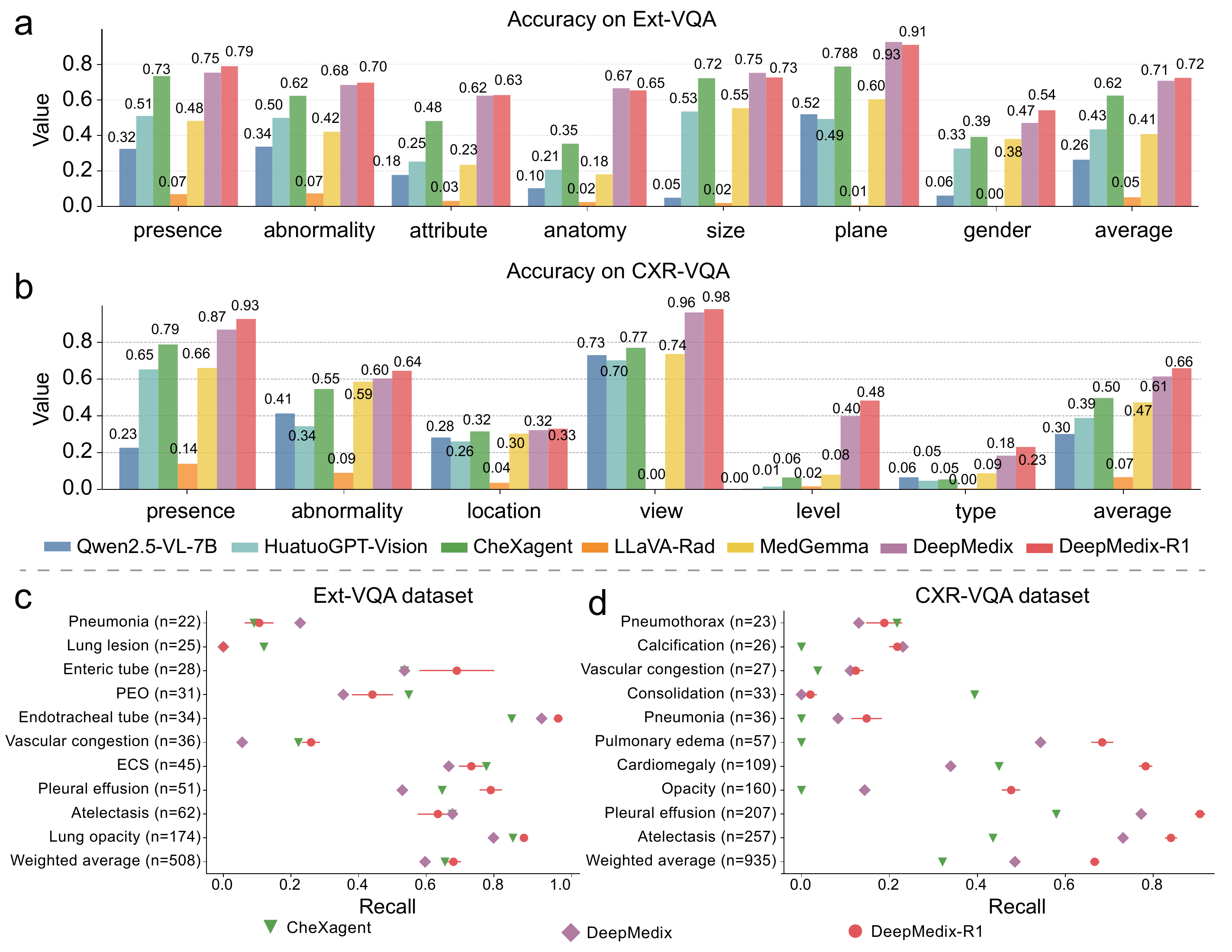}
        \vspace{-0.75em}
    \caption{
    The results of the VQA tasks, demonstrating that our \oursreasoning consistently outperforms competing models from both general and fine-grained perspectives. (a) and (b) are the accuracies across all different question types on Ext-VQA and CXR-VQA, respectively. (c) and (d) are the detailed F1 scores on top-10 observations on Ext-VQA and CXR-VQA, respectively.
     ``PEO'' is short for ``Pulmonary edema/hazy opacity'', and ``ECS'' is short for ``Enlarged cardiac silhouette''.
    }
    \label{fig_vqa}
    \vspace{-0.75em}
\end{figure*}

The detailed results of the VQA are presented in Tab.~\ref{tab_extvqa} and Tab.~\ref{tab_cxrvqa}, with additional visualizations provided in Fig.~\ref{fig_vqa}.
As shown in the results, general-purpose models exhibit suboptimal performance in this domain. On the \extvqa dataset, \oursreasoning demonstrates superior accuracy across multiple question types, including presence, abnormality, attribute, anatomy, size, plane, and gender. Notably, \ours achieves particularly competitive results in anatomy, size, and plane-related questions (e.g., 0.9101 accuracy for plane questions).
On average, \oursreasoning delivers significant performance improvements over state-of-the-art medical FMs, e.g., 66.97\% over \huatuo (0.4338), 16.04\% over \chexagent (0.6242), 77.79\% over \medgemma (0.4074).
Similarly, on the \cxrvqa dataset, \oursreasoning attains the highest accuracy across all question categories (presence, abnormality, location, view, level, and type). For instance, it achieves an accuracy of 0.9800 on ``view'' questions. Compared to leading medical FMs, \oursreasoning shows substantial gains, e.g., 69.75\% over \huatuo (0.3881), 32.52\% over \chexagent (0.4972), 39.27\% over \medgemma (0.4731).
In general, RL enhances VQA performance, as evidenced by improved scores on both datasets (0.7243 vs. 0.7079 and 0.6589 vs. 0.6141, respectively).

We also compute recall rates for key abnormalities, attributes, and anatomical structures, with the top-10 highest-frequency objects visualized in Fig.~\ref{fig_vqa} (c) and (d).
On average of \extvqa dataset, our reasoning model \oursreasoning achieves an average recall of 0.6804, demonstrating significant improvements over medical FMs, e.g., 3.80\% over \chexagent (0.6555) and 14.07\% over \ours (0.5965).
Notably, \oursreasoning performs best on the following objects: ``Lung opacity'', ``Pleural effusion'', ``Vascular congestion'', ``Endotracheal tube'', and ``Enteric tube''.
On average of \cxrvqa dataset, the performance gains are even more pronounced. Compared to medical FMs, \oursreasoning achieves an average recall of 0.6670, with improvements of 107.85\% over \chexagent (0.3209) and 37.36\% over \ours (0.4856).
For nearly all objects,
\oursreasoning delivers the highest performance.

\begin{table*}[t!]
\caption{Medical diagnosis performance on external data, evaluated across seven findings with their respective case counts and averaged results. The assessment employs precision, recall, and F1-score as the evaluation metrics. 
}
\vspace{-0.35em}
\centering
\resizebox{0.99\textwidth}{!}{
\setlength{\tabcolsep}{4pt}
\begin{tabular}{lcccccccccccc}
\toprule
\multicolumn{1}{c}{\multirow{2}{*}{\textbf{Model}}}& \multicolumn{3}{c}{Edema (236)}&	\multicolumn{3}{c}{Pleural thickening (352)}&\multicolumn{3}{c}{Effusion (378)}&\multicolumn{3}{c}{Cardiomegaly (282)} \\
\cmidrule(lr){2-4} \cmidrule(lr){5-7} \cmidrule(lr){8-10} \cmidrule(lr){11-13}
& P&R&F1& P&R&F1& P&R&F1& P&R&F1\\
\midrule
Qwen2.5-VL-7B&0.6364&0.1186&0.2000&0.0000&0.0000&0.0000&0.5283&0.1233&0.2000&0.6207&0.2553&0.3618 \\
Qwen2.5-VL-7B (think)&0.5455&0.1017&0.1714&0.5058&0.4943&0.5000&0.4444&0.0423&0.0773&0.5018&0.9787&0.6635\\
Qwen2.5-VL-32B&0.5616&0.3475&0.4293&0.5079&0.1818&0.2678&0.5119&0.3789&0.4354&0.5537&0.4752&0.5115 \\
DeepSeek-VL-7B&0.4957&0.9661&0.6552&0.4984&0.9091&0.6439&0.5000&0.9824&0.6627&0.4943&0.9291&0.6453 \\
HuatuoGPT-Vision&0.6559&0.5169&0.5782&0.5952&0.1420&0.2294&0.4634&0.1674&0.2460&0.4889&0.1560&0.2366\\
CheXagent&0.6575&0.8136&\underline{0.7273}&0.5050&0.2898&0.3682&0.7173&0.6035&0.6555&0.9024&0.2624&0.4066\\
LLaVA-Rad&0.5108&1.0000&0.6762&0.5029&1.0000&\underline{0.6692}&0.5000&0.9912&0.6647&0.4964&0.9787&0.6587\\
MedGemma&0.6383&0.7627&0.6950&0.6111&0.0625&0.1134&0.6220&0.7974&\underline{0.6988}&0.6514&0.8085&\underline{0.7215} \\
DeepMedix&0.7126&0.5254&0.6049&0.6392&0.3523&0.4542&0.7099&0.4097&0.5196&0.5649&0.9574&0.7105 \\
DeepMedix-R1&0.6294&0.9068&\textbf{0.7431}&0.6008&0.8125&\textbf{0.6908}&0.6574&0.8370&\textbf{0.7364}&0.5887&0.9645&\textbf{0.7312}\\

\midrule

& \multicolumn{3}{c}{Emphysema (254)}&	\multicolumn{3}{c}{Fibrosis (168)}&\multicolumn{3}{c}{Consolidation (452)}&\multicolumn{3}{c}{Average (2122)} \\
\cmidrule(lr){2-4} \cmidrule(lr){5-7} \cmidrule(lr){8-10} \cmidrule(lr){11-13}
& P&R&F1& P&R&F1& P&R&F1& P&R&F1\\
\midrule
Qwen2.5-VL-7B&0.5556&0.2756&0.3684&0.5373&0.4286&0.4768&0.5474&0.5619&0.5546&0.4730&0.2557&0.3059\\
Qwen2.5-VL-7B (think)&0.5227&0.5433&0.5328&0.5217&0.1429&0.2243&0.5462&0.3142&0.3989&0.5106&0.3742&0.3704\\
Qwen2.5-VL-32B&0.2000&0.0079&0.0152&0.5714&0.0952&0.1633&0.6357&0.3938&0.4863&0.5161&0.2918&0.3560\\
DeepSeek-VL-7B&0.5042&0.9370&0.6556&0.5185&1.0000&\underline{0.6829}&0.4978&0.9823&0.6607&0.5000&0.9573&0.6568\\
HuatuoGPT-Vision&0.5982&0.5276&0.5607&0.5068&0.8810&0.6435&0.5179&0.9602&\underline{0.6729}&0.5412&0.4690&0.4390\\
CheXagent&0.8214&0.7244&\textbf{0.7699}&0.5429&0.6786&0.6032&0.6429&0.6770&0.6595&0.6828&0.5656&0.5932 \\
LLaVA-Rad&0.4980&0.9921&0.6632&0.5030&1.0000&0.6693&0.5000&1.0000&0.6667&0.5012&0.9947&\underline{0.6665}  \\
MedGemma&0.6023&0.8346&0.6997&0.5625&0.1071&0.1800&0.6070&0.6150&0.6110&0.6156&0.5841&0.5446  \\
DeepMedix&0.5000&1.0000&0.6667&0.5000&1.0000&0.6667&0.6147&0.6283&0.6214&0.6172&0.6498&0.5945\\
DeepMedix-R1&0.6129&0.8976&\underline{0.7284}&0.6058&0.9881&\textbf{0.7511}&0.6071&0.8274&\textbf{0.7004}&0.6156&0.8748&\textbf{0.7214}\\

\bottomrule
\end{tabular}
}
\label{tab_ext_data}
\vspace{-0.75em}
\end{table*}

\subsection{Medical diagnosis on external data}

We evaluate the medical diagnostic capability of the models on the external CXR14 dataset~\cite{wang2017chestx} by directly posing the question, ``Is there \emph{abnormal X}?'' for each CXR image.
The performance metrics, i.e., precision, recall, and F1-score, are summarized in Tab.~\ref{tab_ext_data}.
The evaluation includes 2,122 cases distributed across seven pathologies: ``Edema'' (236), ``Pleural thickening'' (352), ``Effusion'' (378), ``Cardiomegaly'' (282), ``Emphysema'' (254), ``Fibrosis'' (168), and ``Consolidation'' (452).
Across all pathologies, DeepMedix-R1 achieves the best overall performance, with the highest mean F1-score of 0.7214.
The Qwen-series models, such as \qwenthink and \qwenlarge, exhibit markedly lower mean F1-scores (e.g., 0.3704 and 0.3560), underscoring their limited generalizability to this diagnostic task.
\deepseek and \llavarad achieve notably high recall but relatively low precision, indicating a strong tendency to predict the presence of pathology across most diagnostic queries, regardless of the actual findings.
Notably, CheXagent exhibits the strongest performance in detecting ``Emphysema'', achieving an F1-score of 0.7699, with high precision (0.8214) and recall (0.7244). DeepMedix-R1 followes closely with an F1-score of 0.7284. Overall, this comprehensive evaluation demonstrates that DeepMedix-R1 delivers the most robust and consistent diagnostic performance on external data, as evidenced by its leading average F1-score—representing relative improvements of 21.63\% over CheXagent and 8.24\% over LLaVA-Rad.

\begin{table}[t!]
\caption{Expert annotation results across five dimensions. \emph{Rel.}, \emph{Cor.}, \emph{Comp.}, \emph{GP}, and \emph{OP} are short for relevance, correctness, completeness, grounded preference, and overall preference, respectively. The \emph{agreement} illustrates inter-group labeling consistency.
}
\centering
\resizebox{0.5\textwidth}{!}{
\setlength{\tabcolsep}{4pt}
\begin{tabular}{lc|ccccc}
\toprule
\multicolumn{2}{c|}{\textbf{Model}}& \textbf{Rel.}&	\textbf{Cor.}&	\textbf{Comp.}&	\textbf{\makecell{GP}}&	\textbf{\makecell{OP}} \\
\midrule
\multicolumn{1}{c}{\multirow{3}{*}{\textbf{\makecell{Qwen2.5-VL-7B\\(think)}}}}&Group 1&0.9308 & 0.7622 & 0.6220 & 0.1905 & 0.2619 \\
&Group 2&0.9899 & 0.7441 & 0.9238 & 0.2679 & 0.2548 \\
&Average&0.9603 & 0.7531 & 0.7729 & 0.2292 & 0.2584 \\
\cmidrule(r){1-7}
\multicolumn{1}{c}{\multirow{3}{*}{\textbf{DeepMedix-R1}}}&Group 1& 0.9317 & 0.7014 & 0.7647 & 0.8095 & 0.7381 \\
&Group 2&0.9886 & 0.7993 & 0.9536 & 0.7321 & 0.7452 \\
&Average&0.9601 & 0.7503 & 0.8592 & 0.7708 & 0.7416 \\
\cmidrule(r){1-7}
\multicolumn{2}{c|}{Agreement}&0.9190 & 0.7686 & 0.7447 & 0.7762 & 0.7101 \\
\bottomrule
\end{tabular}
}
\label{tab_annotation}
\vspace{-0.75em}
\end{table}

\subsection{Expert evaluation for reasoning process}
We present two concrete case studies in Fig.~\ref{fig_case1} and \ref{fig_case2} for findings generation and VQA, respectively, with all five annotation dimensions clearly illustrated.
The final annotation of the reasoning process is shown in Tab.~\ref{tab_annotation}.
The agreement of the two groups is relatively high, with the highest 0.9190 at relevance and the lowest 0.7101 at overall preference.
\qwenthink and \oursreasoning perform comparably and are high at \emph{Relevance} (0.9603 vs. 0.9601)
but less at \emph{Correctness} (0.7531 vs. 0.7503), indicating FMs usually generate relevant steps towards the answer while there is a large room for completely correct expression, regarded as a widespread hallucination problem in generative models~\cite{farquhar2024detecting}.
The \emph{Completeness} of \qwenthink is much less than \oursreasoning (0.7729 vs. 0.8592), demonstrating \qwenthink model would miss some key reasoning steps.
For model preference, \oursreasoning scores higher than \qwenthink at both grounded and overall parts (0.7708 vs. 0.2292, 0.7416 vs. 0.2584), indicating our model is more likely to be accepted by professional doctors.

\section{Related Work}

The evolution of LLMs has paved the way for medical FMs by fine-tuning specifically on clinical data.
Notable examples include HuatuoGPT~\cite{zhang2023huatuogpt}, which performs like a doctor in many aspects; LLaVA-Rad~\cite{zambrano2025clinically}, tailored for CXR interpretation; and Dia-LLaMA~\cite{chen2025dia}, specifically optimized for automated CT report generation.

However, these models remain constrained in their ability to generate transparent intermediate reasoning chains that substantiate their final outputs~\cite{wu2025neuro,lin2026survey}.
To address this, recent approaches have adopted GRPO to elicit more robust reasoning behaviors.
enhance the reasoning ability for medical question answering by GRPO.
In the medical domain, Fleming-R1~\cite{liu2025fleming} and Med-U1~\cite{zhang2025med} leverage GRPO to bolster reasoning in medical question-answering tasks.
Med-R1~\cite{lai2025med} employs vanilla GRPO to facilitate medical reasoning across eight distinct imaging modalities.
Furthermore, OphthaReason~\cite{wu2025bridging} represents the first ophthalmology-specific multimodal reasoning model. It introduces an uncertainty-aware dynamic thinking mechanism that utilizes sample-level uncertainty to adaptively regulate the learning process. Despite these advancements, current models still lack fine-grained local understanding and human-comparable reasoning capabilities when processing multimodal medical images.

\section{Conclusion}

We developed \oursreasoning, a FM that integrates grounded reasoning
with online RL for medical image interpretation.
Our three-stage training pipeline, i.e., instruction fine-tuning, cold-start reasoning, and online RL,
represents a significant advance toward transparent, clinically actionable AI.
The experiment results show that \oursreasoning substantially outperforms advanced FMs, achieving improvements in report generation, VQA, and external diagnosis tasks.
Crucially, a formal evaluation by clinical experts demonstrates a clear preference for the reasoning generated by \oursreasoning compared to the widely utilized Qwen2.5-VL-7B model.
In summary, by combining grounded reasoning, our model not only achieves superior performance but also enhances trust and usability in real-world clinical settings.
Thus, our work advances medical FM development toward holistic, transparent, and clinically actionable modeling for CXR interpretation.

\section*{Impact Statement}

This work introduces \oursreasoning, a foundation model for CXR understanding. While effective, it still faces several limitations as detailed in Appendix \S\ref{sec_discussion}. This work poses no ethical concerns, as it relies solely on publicly available datasets and aims to advance medical intelligent applications with positive societal impacts.

\bibliography{example_paper}
\bibliographystyle{icml2026}

\appendix
\onecolumn

\section{Experimental Setups}

\subsection{Experimental Setups}

\begin{figure*}[h]
    \centering
        \includegraphics[width=0.96\linewidth]{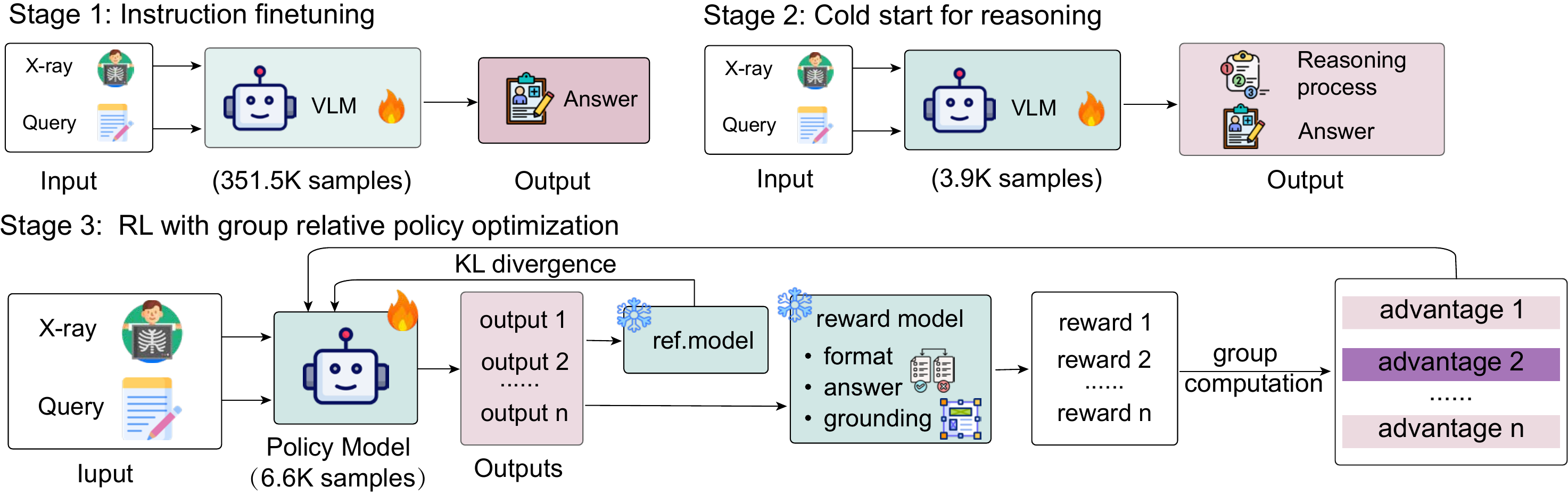}
    \caption{The illustration of the supervised fine-tuning and RL stages, where the icons \snow\; and \fire\; indicate that the model is frozen and updating during training, respectively.
    }
    \label{fig_arc}
\end{figure*}

\begin{figure}[h]
    \centering
        \includegraphics[width=0.5\linewidth]{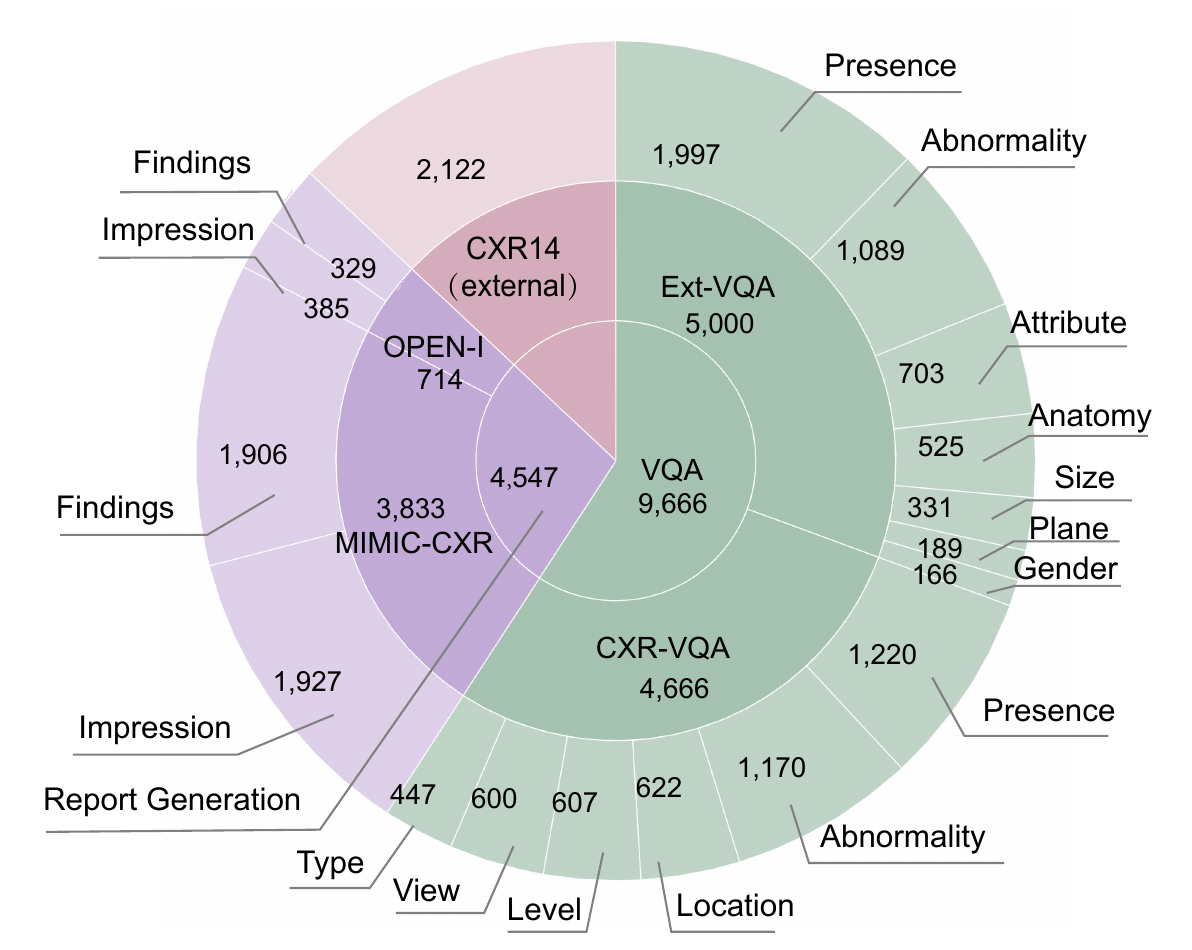}
    \caption{The data distribution of the XrayBench for testing.
    }
    \label{fig_bench}
\end{figure}

We present the overall workflow of \oursreasoning in Fig.~\ref{fig_arc}.
Also, the statistical overview of all utilized datasets is presented in Tab.~\ref{tab_dataset}, and the statistics of XrayBench for testing is shown in Fig.~\ref{fig_bench}.
For stage 1 and stage 2, we use LLaMA-Factory for fine-tuning; the maximum pixels of images are set to 262,144.
The training epochs for stages 1 and 2 are set to 2 and 1, respectively.
The train batch size per device is 1, and the gradient accumulation steps are 2.
For stage 3, we use EasyR1 for optimization.
The total epoch is 15. The number for rollout is 5 with a temperature of 1.0.
The 10\% data is used for validation.
The rollout batch size and valid batch size are 512 and 1024, respectively.
In the experiment, the 4$\times$A100 80G GPU is utilized.
For the evaluation, the temperature is set to 0.6.
The images are resized to 512$\times$512 for training and testing.
Since both \extvqa and \cxrvqa are derived from the MIMIC-CXR dataset, we strictly enforce dataset separation to ensure that all CXR images used in the VQA evaluation are \textbf{completely unseen during model training}.
The experimental prompts and annotation template are illustrated in Fig.~\ref{fig_prompt} and Fig.~\ref{fig_annotation}.

\begin{table*}[h]
\caption{Training and testing data utilized in the paper.
}
\centering
\resizebox{0.62\textwidth}{!}{
\setlength{\tabcolsep}{4pt}
\begin{tabular}{llcccccc|c}
\toprule
\multicolumn{2}{c}{\multirow{2}{*}{\textbf{Stage}}}& \multicolumn{6}{c|}{\textbf{Datasets}}   &\multicolumn{1}{c}{\multirow{2}{*}{\textbf{Total}}}  \\
\cmidrule(r){3-8}
&&\textbf{MIMIC-CXR}&	\textbf{OPEN-I}&	\textbf{Ext-VQA}&	\textbf{CXR-VQA}&\textbf{MS-CXR} &\textbf{\cxrnew}&\\
\midrule
\multicolumn{1}{c}{\multirow{3}{*}{\textbf{Train}}}&Stage 1& 291,074&5,702&28,716&23,956 &2,040&--&351,488\\
&Stage 2&1,600&341&1,000 & 1,000&--&--& 3,941\\
&Stage 3& 2,398 &574&1,823& 1,805 &--&--& 6,600    \\
\hline
\multicolumn{2}{c}{\multirow{1}{*}{\textbf{Test}}}& 3,833 &714&5,000& 4,666 &--&2,122& 16,335    \\
\bottomrule
\end{tabular}
}
\label{tab_dataset}
\end{table*}

\begin{figure*}
    \centering
    \setlength{\fboxrule}{0.85pt}
    \fbox{ \footnotesize
        \parbox{\textwidth}{\scriptsize
        \texttt{\textbf{\mimicfindings}\\
<image> Please act as an experienced radiologist and generate the "FINDINGS" section of an X-ray report based on the provided image(s). Carefully examine the image(s) and describe all observed anatomical structures and abnormalities in a systematic and objective manner.\\
\textbf{\mimicimpression}\\
<image> Please act as an experienced radiologist and generate a succinct "IMPRESSION" section summarizing the most pertinent findings from the imaging study. Focus on key abnormalities and potential diagnoses, ensuring clarity and conciseness.\\
\textbf{\openifindings}\\
This is a chest X-ray study from Indiana University. Please act as an experienced radiologist and generate the "FINDINGS" section of an X-ray report based on the provided image(s). Carefully examine the image(s) and describe all observed anatomical structures and abnormalities in a systematic and objective manner. The {view} view of the X-ray is <image>\\
\textbf{\openiimporession}\\
This is a chest X-ray study from Indiana University. Please act as an experienced radiologist and generate a succinct "IMPRESSION" section summarizing the most pertinent findings from the imaging study. Focus on key abnormalities and potential diagnoses, ensuring clarity and conciseness. The \{view\} view of the X-ray is <image>\\
\textbf{\extvqa}\\
You are an expert in chest X-ray interpretation. Please answer the following question about the given chest X-ray image. The question is: \{question\}. The X-ray image is <image>\\
\textbf{\cxrvqa}\\
You are an expert in chest X-ray interpretation. Please answer the following question about the given chest X-ray image. The question is: \{question\}. The X-ray image is <image>\\
\textbf{Localization prompt 1}\\
<image> You are an expert in chest X-ray interpretation. Please identify the pathology present in the provided chest X-ray image, along with coordinates (in the format [x\_min, y\_min, x\_max, y\_max]) of the corresponding bounding box and a detailed textual annotation describing the findings. Please output in JSON format.\\
\textbf{Localization prompt 2}\\
<image> You are an expert in chest X-ray interpretation. A bounding box (in the format [x\_min, y\_min, x\_max, y\_max]) is visible within a provided chest X-ray image. Please identify the corresponding pathology for the region enclosed by the bounding box in the chest X-ray. The bounding box coordinates are: \{bounding box\}.\\
\textbf{Think prompt}\\
You FIRST think about the reasoning process as an internal monologue and then provide the final answer. The reasoning process MUST BE enclosed within <think> </think> tags. During this reasoning process, prioritize analyzing the local regions of the image by leveraging the bounding box coordinates in the format [x\_min, y\_min, x\_max, y\_max]. The final answer MUST BE put in \textbackslash boxed\{\}. An example is like: <think> reasoning process 1 with [x\_min1, y\_min1, x\_max1, y\_max1]; reasoning process 2 with [x\_min2, y\_min2, x\_max2, y\_max2] </think>. The answer is: \textbackslash boxed\{answer\}.\\
\textbf{Reasoning generation prompt}\\
\# Objective:\\
Generate the reasoning process for the answer, which should include several reasoning steps (ranging from 1 to 5). For each reasoning step, focus on analyzing the local regions of the image by utilizing the bounding box coordinates provided in the format [x\_min, y\_min, x\_max, y\_max].\\
\\
\#\# Guidance:\\
The reasoning process should be structured as follows:\\
1) Reasoning text 1 with [x\_min1, y\_min1, x\_max1, y\_max1];\\
2) Reasoning text 2 with [x\_min2, y\_min2, x\_max2, y\_max2];\\
...
\\
Continue this pattern for each reasoning step as needed.\\
\textbf{Filter prompt}\\
\# Objective: Evaluate the correctness and quality of reasoning in X-ray image analysis based on the given task instruction, answer, and reasoning process. Focus on the following criteria:\\
1) Content Accuracy: Ensure that the content of the reasoning process is factually correct and aligns with established medical knowledge and standards for X-ray interpretation.\\
2) Hallucination-Free Descriptions: Verify that the reasoning process does not include any descriptions that are not supported by the actual X-ray image or medical evidence.\\
3) Conciseness \& Relevance: Confirm that the reasoning process is concise and free from unnecessary or irrelevant information.\\
4) Accuracy of Local Regions: Evaluate whether the reasoning process accurately references specific regions of the X-ray image using bounding box coordinates.\\\\
\#\# Guidance:\\
- The evaluation levels are categorized into three distinct categories: Qualified, Possibly Qualified, Unqualified.\\
- The result label should be enclosed in the format: <Evaluation> ... </Evaluation>\\
For example, an evaluation result may appear as follows:<Evaluation> Possibly Qualified </Evaluation>
        }
    }}
    \caption{Prompts in the experiments.}
    \label{fig_prompt}
\end{figure*}

\begin{figure*}
    \centering
    \setlength{\fboxrule}{0.85pt}
    \fbox{ \footnotesize
        \parbox{0.85\textwidth}{\texttt{\textbf{Annotation sample No: \{number\}}\\
\textbf{Task: \{Task prompt\}}\\
\textbf{Answer: \{Ground truth\}}\\
\\
\\
\textbf{Output from model 1:}\\
\textbf{Answer from model 1: \{answer1\}}\\
\begin{minipage}{\textwidth}
            \begin{tabular}{|l|l|l|}
                \toprule
                \textbf{Reasoning steps} & \textbf{relvant (1/0)}&\textbf{correct (1/0)} \\
                \midrule
                \emph{Step 1 with details} &\faSquare &\faSquare \\
                \hline
                \emph{Step 2 with details} &\faSquare &\faSquare \\
                \hline
                \emph{...} & \faSquare &\faSquare \\
                \hline
                \emph{Step n with details} &\faSquare & \faSquare\\
                \bottomrule
            \end{tabular}
\end{minipage}\\
\\
\begin{minipage}{\textwidth}
            \begin{tabular}{|l|p{1cm}|}
                \toprule
                \textbf{Whether this reasoning process is complete (1/0)} &\faSquare  \\
                \bottomrule
            \end{tabular}
\end{minipage}\\
\\
\\
\textbf{Output from model 2:}\\
\textbf{Answer from model 2: \{answer2\}}\\
\begin{minipage}{\textwidth}
            \begin{tabular}{|l|l|l|}
                \toprule
                \textbf{Reasoning steps} & \textbf{relvant (1/0)}&\textbf{correct (1/0)} \\
                \midrule
                \emph{Step 1 with details} &\faSquare &\faSquare \\
                \hline
                \emph{Step 2 with details} &\faSquare &\faSquare \\
                \hline
                \emph{...} &\faSquare &\faSquare \\
                \hline
                \emph{Step n with details} &\faSquare &\faSquare \\
                \bottomrule
            \end{tabular}
\end{minipage}\\
\\
\begin{minipage}{\textwidth}
            \begin{tabular}{|l|p{1cm}|}
                \toprule
                \textbf{Whether this reasoning process is complete (1/0)} & \faSquare \\
                \bottomrule
            \end{tabular}
\end{minipage}\\
\\
\\
\begin{minipage}{\textwidth}
            \begin{tabular}{|l|p{1cm}|}
                \toprule
                \textbf{Grounded expression preference (model 1/model 2)} & \faSquare \\
                \hline
                \textbf{Overall reasoning process preference (model 1/model 2)} & \faSquare \\
                \bottomrule
            \end{tabular}
\end{minipage}\\
        }
    }}
    \caption{Annotation template. The \faSquare \;symbol denotes the regions requiring expert annotation.}
    \label{fig_annotation}
\end{figure*}


\subsection{Evaluation}

The evaluation of medical FMs presents a critical yet formidable challenge.
Unlike conventional AI systems that produce structured outputs~\cite{lin2025self,yan2026tide,xu2026odysseyarena}, medical FMs typically generate unstructured free-text responses (e.g., diagnostic reports), making automated assessment complicated~\cite{lin2025cross,hou2025improving}. This challenge is compounded by the fact that divergent responses may represent clinically plausible variations rather than errors, necessitating expert evaluation with specialized medical knowledge. Consequently, this reliance on human expertise introduces inherent subjectivity and scalability limitations~\cite{chang2024survey}.
Conventional natural language generation (NLG) metrics (e.g., BLEU~\cite{DBLP:conf/acl/PapineniRWZ02}, METEOR~\cite{DBLP:conf/wmt/LavieA07}, and ROUGE-L~\cite{lin2004rouge}) are fundamentally limited to assessing surface-level textual similarity and fail to capture clinical validity and nuance, highlighting the need for domain-specific evaluation frameworks. To this end, we introduce Report Arena for in-depth model comparison.

\subsection{Data availability}
The MIMIC-CXR dataset is publicly accessible on PhysioNet (https://physionet.org/content/mimic-cxr-jpg/2.1.0/).
The OpenI dataset can be downloaded from Kaggle (https://www.kaggle.com/datasets/raddar/chest-xrays-indiana-university).
The MS-CXR dataset is available on PhysioNet (https://physionet.org/content/ms-cxr/1.1.0/).
For visual question-answering (VQA) tasks, the Ext-VQA and CXR-VQA datasets are also available on PhysioNet (https://physionet.org/content/mimic-ext-mimic-cxr-vqa/1.0.0/, https://physionet.org/content/medical-cxr-vqa-dataset/1.0.0/)
The CXR14 is available via Kaggle (https://www.kaggle.com/datasets/nih-chest-xrays/sample).


\subsection{Datasets and Baselines}
The datasets employed for this study are detailed below:
\begin{itemize}
    \item MIMIC-CXR~\cite{johnson2019mimic} is one of the most widely used chest CXR datasets, containing 377,110 images in JPG format with structured labels derived from 227,827 corresponding free-text radiology reports. This dataset is employed for the report generation task in both the training and evaluation phases.
    \item Open-I~\cite{demner2012design} is a widely used open-access chest X-ray dataset from Indiana University, comprising 7,470 images and corresponding radiology reports.
    \item \extvqa~\cite{bae2024mimic} a large-scale, diverse, and complex dataset designed for VQA tasks in the chest X-ray (CXR) domain, built upon the MIMIC-CXR-JPG, MIMIC-IV, and Chest ImaGenome datasets. To enhance linguistic diversity while preserving medical relevance, the dataset employs a paraphrasing strategy that generates an average of 16.5 paraphrases per template using carefully crafted GPT-4–based prompts.
    \item \cxrvqa~\cite{Hu2025cxrvqa} finetunes a LLM to build an intermediate KeyInfo dataset, from which question–answer pairs are subsequently extracted to construct the final dataset, based on MIMIC-CXR. For computational efficiency, we use subsets of both \extvqa and \cxrvqa. 
    \item MS-CXR~\cite{boecking2022ms} comprises 1,162 image–sentence pairs that associate bounding boxes with descriptive phrases, spanning eight distinct radiological findings with roughly balanced representation across categories. We employ this dataset during training stage 1 to endow the model with visual–textual grounding capability.
    \item \cxrnew~\cite{wang2017chestx} is a widely used CXR diagnostic dataset containing a substantial collection of frontal-view radiographs annotated with multiple clinical findings. For evaluation, we utilize 2,122 samples as the test set.
\end{itemize}

To assess the performance of our proposed model, we benchmark it against the most powerful and widely adopted VLMs, encompassing both general-purpose and domain-specific categories:

\begin{itemize}
    \item \qwen~\cite{qwen2.5-VL} is part of the Qwen2.5-VL series, a suite of advanced general-purpose VLMs developed by Alibaba Cloud, capable of processing and understanding diverse modalities including text, images, and videos. We also include \emph{Qwen2.5-VL-7B (think)}, which incorporates specialized prompting to enable the model to generate both the reasoning process and the final answer simultaneously.
    \item \qwenlarge~\cite{qwen2.5-VL} is similar to \qwen and is also part of the Qwen2.5-VL series. It contains more parameters and consequently demonstrates stronger overall capability.
    \item \deepseek~\cite{lu2024deepseek} is designed for real-world multimodal understanding tasks. It is built upon the DeepSeek-LLM-7B-base and DeepSeek-LLM-7b-base and employs a hybrid vision encoder composed of SigLIP-L\footnote{https://huggingface.co/timm/ViT-L-16-SigLIP-384} and SAM-B\footnote{https://huggingface.co/facebook/sam-vit-base}.
    \item HuatuoGPT-Vision~\cite{chen2024huatuogptvisioninjectingmedicalvisual} is a VLM tailored for medical applications, building upon the Qwen2.5-VL-7B model via training on the PubMedVision dataset.
    \item MedGemma~\cite{medgemma-hf}, developed by Google, is an FM tailored for medical applications, and we use its 4-billion-parameter version for comparison. The model is based on Gemma 3 and integrates a SigLIP image encoder pre-trained on a diverse collection of medical images, including CXRs and histopathology slides.
    \item CheXagent~\cite{chen2024chexagent} is designed for comprehensive interpretation of CXRs and is trained on the CheXinstruct comprising 8.5 million instruction-image–response triplets. We use its 8-billion-parameter version as the default model.
    \item LLaVA-Rad~\cite{zambrano2025clinically} is a 7-billion-parameter multimodal model designed to generate diagnostic findings from input CXRs. Its architecture builds upon that of LLaVA~\cite{liu2023visual}, training on a large-scale dataset comprising over 697K image–text pairs.
\end{itemize}

\section{Appendix for Experimental Results}

\subsection{Length Distribution}

\begin{figure*}[h]
    \centering
        \includegraphics[width=0.96\linewidth]{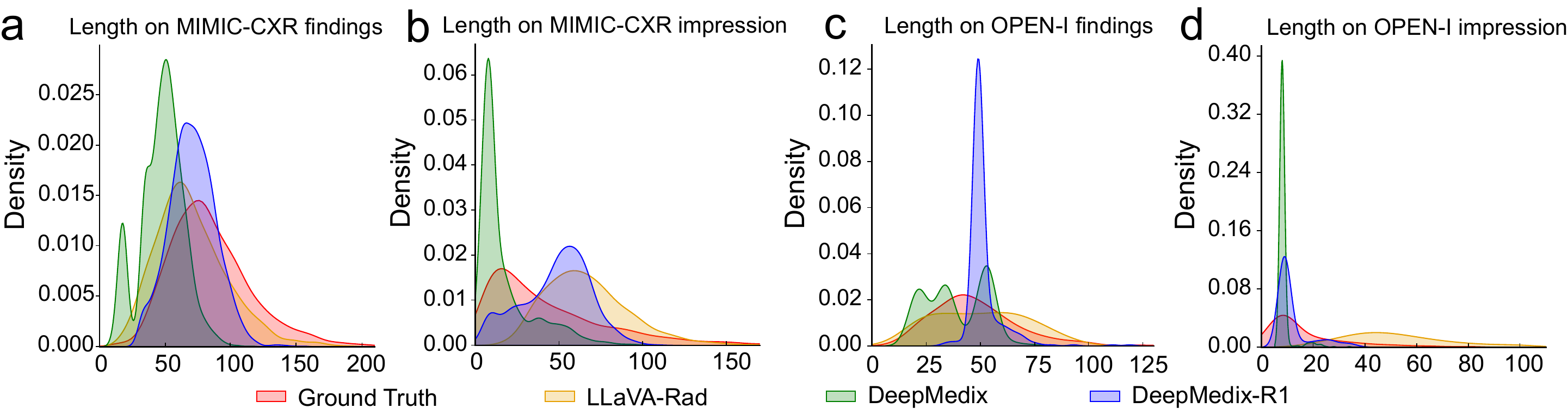}
    \caption{
    The length distributions of the generated report in \mimicfindings, \mimicimpression, \openifindings, and \openiimporession, respectively.
    }
    \label{fig_length}
\end{figure*}

At the beginning, to have an overview of the model generation, we provide length distributions for ground truth and results from \llavarad, \ours, and \oursreasoning, on \mimicfindings, \mimicimpression, \openifindings, and \openiimporession.
The results are shown in Fig.~\ref{fig_length} (a), (b), (c), and (d), respectively.
From them, we find that after the first stage, \ours and \llavarad demonstrate comparable results to the ground truth in terms of report length alignment.
\llavarad wound generate more tokens on \mimicimpression and \openifindings than ground truth, which could potentially produce unnecessary or incorrect information.
After online RL, the distribution of model \oursreasoning demonstrates a more pronounced shift toward the desired direction to truth on \mimicfindings and \openiimporession, compared with \ours.
On \mimicimpression and \openifindings, \oursreasoning would generate more tokens and be more inclined to generate responses of average length (i.e., 40--60).

\subsection{Detailed Experimental Results}

The comprehensive experimental results for report generation, \extvqa, and \cxrvqa tasks are presented in Tab.~\ref{tab_rg}, Tab.~\ref{tab_extvqa}, and Tab.~\ref{tab_cxrvqa}, respectively.

\begin{table*}[t!]
\caption{The report generation results are presented for \mimicfindings, \mimicimpression, \openifindings, and \openiimporession, followed by the weighted average across these four splits.
The metrics Micro-5, Macro-5, Micro-14, and Macro-14 are obtained from F1-CheXbert.
The \textbf{best} and \underline{second}-best values for each metric are highlighted in bold and boxed, respectively.
MTR, R-L, and F1-R are short for METEOR, ROUGE-L, and F1-RadGraph, respectively.
}
\centering
\resizebox{0.99\textwidth}{!}{
\setlength{\tabcolsep}{4pt}
\begin{tabular}{lcccccccccccc}
\toprule
\multicolumn{1}{c}{\multirow{2}{*}{\textbf{Model}}}& \multicolumn{6}{c}{\textbf{NLG-based Metric}}   & \multicolumn{5}{c}{\textbf{Fact-based Metric}}&\multicolumn{1}{c}{\multirow{2}{*}{\textbf{Average}}}  \\
\cmidrule(r){2-7} \cmidrule(r){8-12}
\multicolumn{1}{c}{}& \textbf{BLEU-1}  & \textbf{BLEU-2} & \textbf{BLEU-3} & \textbf{BLEU-4} & \textbf{MTR}  & \textbf{R-L} & \textbf{F1-R} & \textbf{Macro-14}&\textbf{Micro-14}&\textbf{Macro-5}&\textbf{Micro-5}&  \\
\midrule
\multicolumn{13}{c}{\cellcolor{gray!50}MIMIC-CXR Findings}\\
Qwen2.5-VL-7B& 0.0732 & 0.0367 & 0.0200 & 0.0117 & 0.1932 & 0.1434   & 0.0930     & 0.1327       & 0.2137       & 0.1643      & 0.1769         & 0.1144         \\
\qwenthink & 0.0615 & 0.0304 & 0.0197 & 0.0147 & 0.0764 & 0.1106   & 0.0493     & 0.1198       & 0.1741       & 0.1492      & 0.1509         & 0.0870         \\
Qwen2.5-VL-32B& 0.0887 & 0.0440 & 0.0235 & 0.0136 & 0.2127 & 0.1538   & 0.0921     & 0.2260       & 0.3461       & 0.3258      & 0.3476         & 0.1704         \\
DeepSeek-VL-7B& 0.1356 & 0.0566 & 0.0300 & 0.0194 & 0.1673 & 0.1730   & 0.0619     & 0.1300       & 0.2444       & 0.1118      & 0.1710         & 0.1183         \\
HuatuoGPT-Vision& 0.2244 & 0.1133 & 0.0631 & 0.0390 & 0.2361 & 0.1998   & 0.1127     & 0.2222       & 0.3532       & 0.3543      & 0.4099         & 0.2116        \\
CheXagent& 0.0277 & 0.0170 & 0.0134 & 0.0115 & 0.0424 & 0.0658   & 0.0965     & 0.1275       & 0.1818       & 0.1935      & 0.2479         & 0.0932         \\
LLaVA-Rad& \underline{0.2911} & \underline{0.1757} & \underline{0.1189} & 0.0846 & 0.2500 & \underline{0.2928}   & \underline{0.2077}     & \textbf{0.3154}       & 0.4753       & 0.4592      & \underline{0.5535}         & \underline{0.2931}         \\
MedGemma& 0.1752 & 0.0960 & 0.0586 & 0.0377 & \underline{0.2613} & 0.2122   & 0.1864     & 0.3347       & \underline{0.5041}       & \underline{0.4616}      & 0.5183         & 0.2587         \\
DeepMedix& 0.2445 & 0.1589 & 0.1147 & \underline{0.0871} & 0.2291 & 0.2870   & 0.1832     & 0.2271       & 0.3956       & 0.3476      & 0.4477         & 0.2475         \\
DeepMedix-R1& \textbf{0.3402} & \textbf{0.2149} & \textbf{0.1479} & \textbf{0.1059} & \textbf{0.2895} & \textbf{0.3292}   & \textbf{0.2387}     & \underline{0.3126}       & \textbf{0.5189}       & \textbf{0.4821}      & \textbf{0.5726}         & \textbf{0.3230}         \\
\midrule
\multicolumn{13}{c}{\cellcolor{gray!50}MIMIC-CXR Impression}\\
Qwen2.5-VL-7B& 0.0616 & 0.0280 & 0.0164 & 0.0112 & 0.1323 & 0.0975   & 0.0710     & 0.2072       & 0.3053       & 0.3258      & 0.3635          & 0.1473          \\
\qwenthink & 0.0628 & 0.0414 & 0.0347 & 0.0311 & 0.0720 & 0.0836   & 0.0475     & 0.1475       & 0.2199       & 0.2088      & 0.2352          & 0.1077          \\
Qwen2.5-VL-32B& 0.0624 & 0.0279 & 0.0164 & 0.0113 & 0.1392 & 0.0939   & 0.0585     & 0.2028       & 0.3117       & 0.3190      & 0.3454          & 0.1444          \\
DeepSeek-VL-7B& 0.0624 & 0.0272 & 0.0171 & 0.0126 & 0.1026 & 0.0942   & 0.0273     & 0.1386       & 0.2245       & 0.1288      & 0.1634          & 0.0908          \\
HuatuoGPT-Vision& 0.1176 & 0.0556 & 0.0342 & 0.0243 & 0.1651 & 0.1225   & 0.0714     & 0.1959       & 0.3227       & 0.3347      & 0.3874          & 0.1665         \\
CheXagent& 0.1795 & 0.1337 & \underline{0.1140} & \underline{0.1033} & 0.2032 & \underline{0.2162}   & \textbf{0.1792}     & 0.2359& 	0.4130 &	0.3866& 	0.4417& 	\underline{0.2369}  \\
LLaVA-Rad& \underline{0.1852} & 0.1074 & 0.0741 & 0.0546 & \underline{0.2135} & 0.1961   & 0.1396     & \underline{0.2498}       & \underline{0.4258}       & 0.4195& 	\underline{0.4865}& 	0.2320           \\
MedGemma& 0.1321 & 0.0699 & 0.0480 & 0.0361 & 0.1759 & 0.1368   & 0.1177     & 0.2465 &	0.4146 &	\underline{0.4199}& 	0.4861 &	0.2076  \\
DeepMedix& 0.1682 & \textbf{0.1378} & \textbf{0.1231} & \textbf{0.1151} & 0.1884 & 0.2089   & \underline{0.1779}     & 0.2100       & 0.3744       & 0.3346      & 0.4314          & 0.2245          \\
DeepMedix-R1& \textbf{0.2135} & \underline{0.1376} & 0.1035 & 0.0832 & \textbf{0.2379} & \textbf{0.2191}   & 0.1739     & \textbf{0.2697}       & \textbf{0.4491}       & \textbf{0.4249}      & \textbf{0.5071}          & \textbf{0.2563}          \\
\midrule
\multicolumn{13}{c}{\cellcolor{gray!50}OPEN-I Findings}\\
Qwen2.5-VL-7B& 0.0468 & 0.0263 & 0.0165 & 0.0104 & 0.1706 & 0.1298   & 0.1344     & 0.0576       & 0.1302       & 0.0264      & 0.0280          & 0.0706          \\
\qwenthink & 0.0984 & 0.0575 & 0.0418 & 0.0336 & 0.1000 & 0.1418   & 0.0939     & 0.0729       & 0.3281       & 0.0372      & 0.0451          & 0.0955          \\
Qwen2.5-VL-32B& 0.0577 & 0.0310 & 0.0188 & 0.0118 & 0.1925 & 0.1384   & 0.1083     & 0.0693       & 0.1093       & 0.0631      & 0.0800          & 0.0800          \\
DeepSeek-VL-7B& 0.0822 & 0.0369 & 0.0224 & 0.0154 & 0.1680 & 0.1575   & 0.1366     & 0.0816       & 0.2331       & 0.0573      & 0.1111          & 0.1002          \\
HuatuoGPT-Vision& 0.0883 & 0.0464 & 0.0284 & 0.0184 & 0.2298 & 0.1610   & 0.1483     & 0.1019       & 0.1853       & 0.1524      & 0.2093          & 0.1245         \\
CheXagent& 0.0560 & 0.0376 & 0.0287 & 0.0236 & 0.0643 & 0.0758   & 0.0655     & 0.1364       & 0.2349       & 0.1709      & 0.2750          & 0.1062          \\
LLaVA-Rad& 0.3397 & \underline{0.2150} & 0.1469 & 0.1106 & \underline{0.3032} & \underline{0.3464}   & 0.2859     & 0.2131    & 0.3790 &	\underline{0.2600} &	\underline{0.3211}& 	\underline{0.2655}           \\
MedGemma& 0.1530 & 0.0955 & 0.0643 & 0.0442 & 0.2909 & 0.2150   & 0.2874     & \underline{0.2195}       & 0.3756       & 0.1621      & 0.3143          & 0.2020          \\
DeepMedix& \underline{0.3298} & 0.2098 & \underline{0.1556} & \underline{0.1215} & 0.2916 & 0.3384   & \underline{0.2907}     & 0.1631       & \underline{0.3852}       & 0.1943      & 0.2453          & 0.2478          \\
DeepMedix-R1& \textbf{0.3958} & \textbf{0.2535} & \textbf{0.1834} & \textbf{0.1362} & \textbf{0.3846} & \textbf{0.3753}   & \textbf{0.3267}     & \textbf{0.2498}       & \textbf{0.3905}       & \textbf{0.2952}      & \textbf{0.3286}          & \textbf{0.3018}          \\
\midrule
\multicolumn{13}{c}{\cellcolor{gray!50}OPEN-I Impression}\\
Qwen2.5-VL-7B& 0.0280 & 0.0155 & 0.0111 & 0.0085 & 0.0943 & 0.0590   & 0.0415     & 0.1005       & 0.1826       & 0.1089      & 0.1385          & 0.0717          \\
\qwenthink & 0.0556 & 0.0451 & 0.0414 & 0.0396 & 0.0641 & 0.0614   & 0.0321     & 0.0685       & 0.2162       & 0.0514      & 0.0752          & 0.0682          \\
Qwen2.5-VL-32B& 0.0276 & 0.0147 & 0.0105 & 0.0080 & 0.1002 & 0.0546   & 0.0418     & 0.0790       & 0.1031       & 0.0947      & 0.1044          & 0.0581          \\
DeepSeek-VL-7B& 0.0553 & 0.0289 & 0.0228 & 0.0195 & 0.1117 & 0.0761   & 0.0293     & 0.0910       & 0.4715       & 0.0380      & 0.0561          & 0.0909          \\
HuatuoGPT-Vision & 0.0224 & 0.0111 & 0.0080 & 0.0062 & 0.0781 & 0.0491   & 0.0272     & 0.0751       & 0.0949       & 0.1065      & 0.1296          & 0.0553\\
CheXagent& 0.2544 & 0.2136 & 0.1919 & 0.1764 & 0.2470 & 0.2542   & 0.1815     & 0.1935       & \underline{0.5775}       & 0.1521      & 0.1831          & 0.2387          \\
LLaVA-Rad& 0.1442 & 0.0854 & 0.0647 & 0.0532 & 0.2027 & 0.1450   & 0.1061     & \underline{0.2019}       & 0.5172       & 0.1923      & \underline{0.3333}          & 0.1860          \\
MedGemma& 0.0863 & 0.0533 & 0.0411 & 0.0339 & 0.1566 & 0.1003   & 0.0732     & 0.1910       & 0.4086       & \underline{0.1999}      & 0.2887          & 0.1484          \\
DeepMedix& \textbf{0.3639} & \underline{0.2820} &	\underline{0.2332} &	\underline{0.2305} &	\underline{0.3565}& 	\underline{0.3224}& 	\textbf{0.2909}& 	0.1162 &	0.5847 &	0.1633& 	0.1913& 	\underline{0.2850}   \\
DeepMedix-R1& \underline{0.3426} & \textbf{0.2888} & \textbf{0.2577} & \textbf{0.2377} & \textbf{0.3628} & \textbf{0.3457}   & \underline{0.2893}     & \textbf{0.2061}       & \textbf{0.5915}       & \textbf{0.2837}      & \textbf{0.3774}          & \textbf{0.3258}          \\
\midrule
\multicolumn{13}{c}{\cellcolor{gray!50}Weighted Average}\\
Qwen2.5-VL-7B& 0.0625 & 0.0305 & 0.0175 & 0.0111 & 0.1574 & 0.1158   & 0.0823     & 0.1561       & 0.2438       & 0.2181      & 0.2420         & 0.1216         \\
\qwenthink & 0.0642 & 0.0383 & 0.0295 & 0.0251 & 0.0752 & 0.0972   & 0.0503     & 0.1238       & 0.2082       & 0.1581      & 0.1726         & 0.0948         \\
Qwen2.5-VL-32B& 0.0701 & 0.0338 & 0.0191 & 0.0120 & 0.1706 & 0.1189   & 0.0748     & 0.1924       & 0.2938       & 0.2843      & 0.3067         & 0.1433         \\
DeepSeek-VL-7B& 0.0939 & 0.0404 & 0.0234 & 0.0162 & 0.1352 & 0.1303   & 0.0499     & 0.1268       & 0.2544       & 0.1088      & 0.1537         & 0.1030         \\
HuatuoGPT-Vision& 0.1522 & 0.0754 & 0.0437 & 0.0285 & 0.1922 & 0.1515   & 0.0905     & 0.1899       & 0.3063       & 0.3104      & 0.3621         & 0.1730        \\
CheXagent& 0.1133 & 0.0846 & 0.0723 & 0.0652 & 0.1295 & 0.1462   & 0.1365     & 0.1797       & 0.3171       & 0.2702      & 0.3265         & 0.1674         \\
LLaVA-Rad& \underline{0.2373} & 0.1420 & 0.0974 & 0.0711 & \underline{0.2344} & 0.2432   & 0.1759     & 0.2706       & \underline{0.4509}       & \underline{0.4054}      & \underline{0.4896}         & \underline{0.2562}         \\
MedGemma& 0.1478 & 0.0813 & 0.0530 & 0.0372 & 0.2184 & 0.1710   & 0.1550     & \underline{0.2768}       & 0.4488       & 0.4001      & 0.4705         & 0.2236         \\
DeepMedix& 0.2284 & \underline{0.1641} & \underline{0.1313} & \textbf{0.1136} & 0.2272 & \underline{0.2606}   & \underline{0.1979}     & 0.2058       & 0.4019       & 0.3154      & 0.4044         & 0.2410         \\
DeepMedix-R1& \textbf{0.2907} & \textbf{0.1912} & \textbf{0.1409} & \underline{0.1096} & \textbf{0.2807} & \textbf{0.2873}   & \textbf{0.2219}     & \textbf{0.2809}       & \textbf{0.4862}       & \textbf{0.4275}      & \textbf{0.5107}         & \textbf{0.2934}         \\
\bottomrule
\end{tabular}
}
\label{tab_rg}
\end{table*}

\begin{table*}[h]
\caption{Experimental results on \extvqa dataset.
}
\centering
\resizebox{0.69\textwidth}{!}{
\setlength{\tabcolsep}{4pt}
\begin{tabular}{lccccccc|c}
\toprule
\multicolumn{1}{c}{\multirow{2}{*}{\textbf{Model}}}& \multicolumn{7}{c|}{\textbf{Question types}}   &\multicolumn{1}{c}{\multirow{2}{*}{\textbf{Avg.}}}  \\
\cmidrule(r){2-8}
\multicolumn{1}{c}{}& \textbf{presence}&	\textbf{abnormality}&	\textbf{attribute}&	\textbf{anatomy}&	\textbf{size}&	\textbf{plane}&	\textbf{gender} & \\
\midrule
Qwen2.5-VL-7B& 0.3240    & 0.3361      & 0.1764    & 0.1010   & 0.0483   & 0.5185   & 0.0602   & 0.2628 \\
\qwenthink & 0.5253    & 0.3682      & 0.1508    & 0.0762   & 0.5045   & 0.5714   & 0.2048   & 0.3810 \\
Qwen2.5-VL-32B& 0.4892    & 0.3526      & 0.1451    & 0.0914   & 0.5287   & 0.5873   & 0.1988   & 0.3660 \\
DeepSeek-VL-7B& 0.3380    & 0.2388      & 0.1010    & 0.0724   & 0.1178   & 0.2222   & 0.3072   & 0.2352 \\
HuatuoGPT-Vision& 0.5093    & 0.4986      & 0.2518    & 0.2057   & 0.5347   & 0.4921   & 0.3253   & 0.4338 \\
CheXagent& 0.7346    & 0.6226      & 0.4808    & 0.3524   & 0.7221   & 0.7884   & 0.3916   & 0.6242 \\
LLaVA-Rad& 0.0681    & 0.0725      & 0.0299    & 0.0229   & 0.0181   & 0.0053   & 0.0000   & 0.0510 \\
MedGemma& 0.4812    & 0.4206      & 0.2333    & 0.1790   & 0.5529   & 0.6032   & 0.3795   & 0.4074 \\
DeepMedix& \underline{0.7536}    & \underline{0.6839}      & \underline{0.6235}    & \textbf{0.6658}   & \textbf{0.7523}   & \textbf{0.9259}   & \underline{0.4699}   & \underline{0.7079} \\
DeepMedix-R1& \textbf{0.7894}    & \textbf{0.6970}      & \textbf{0.6269}    & \textbf{0.6532}   & \underline{0.7261}   & \underline{0.9101}   & \textbf{0.5422}   & \textbf{0.7243}\\
\bottomrule
\end{tabular}
}
\label{tab_extvqa}
\end{table*}

\begin{table*}[h]
\caption{Experimental results on \cxrvqa dataset.
}
\centering
\resizebox{0.62\textwidth}{!}{
\setlength{\tabcolsep}{4pt}
\begin{tabular}{lcccccc|c}
\toprule
\multicolumn{1}{c}{\multirow{2}{*}{\textbf{Model}}}& \multicolumn{6}{c|}{\textbf{Question types}}   &\multicolumn{1}{c}{\multirow{2}{*}{\textbf{Avg.}}}  \\
\cmidrule(r){2-7}
\multicolumn{1}{c}{}& \textbf{presence}&	\textbf{abnormality}&	\textbf{location}&	\textbf{view}&	\textbf{level}&	\textbf{type}& \\

\midrule
Qwen2.5-VL-7B& 0.2262    & 0.4128      & 0.2814   & 0.7300   & 0.0033   & 0.0649   & 0.3007 \\
\qwenthink & 0.5885    & 0.2197      & 0.2203   & 0.6200   & 0.0132   & 0.0157   & 0.3213 \\
Qwen2.5-VL-32B& 0.5451    & 0.2966      & 0.1720   & 0.6950   & 0.0445   & 0.0380   & 0.3386 \\
DeepSeek-VL-7B& 0.1418    & 0.1051      & 0.1495   & 0.1500   & 0.0297   & 0.0425   & 0.1106 \\
HuatuoGPT-Vision& 0.6525    & 0.3436      & 0.2605   & 0.7017   & 0.0148   & 0.0470   & 0.3881 \\
CheXagent& 0.7877    & 0.5453      & 0.3151   & 0.7700   & 0.0643   & 0.0537   & 0.4972 \\
LLaVA-Rad& 0.1393    & 0.0895      & 0.0355   & 0.0000   & 0.0164   & 0.0044   & 0.0662 \\
MedGemma& 0.6598    & 0.5855      & 0.3030   & 0.7367   & 0.0791   & 0.0872   & 0.4731 \\
DeepMedix& \underline{0.8689}    & \underline{0.6028}      & \underline{0.3215}   & \underline{0.9617}   & \underline{0.3970}   & \underline{0.1834}   & \underline{0.6141} \\
DeepMedix-R1& \textbf{0.9268}    & \textbf{0.6447}      & \textbf{0.3301}   & \textbf{0.9800}   & \textbf{0.4827}   & \textbf{0.2304}   & \textbf{0.6589}\\
\bottomrule
\end{tabular}
}
\label{tab_cxrvqa}
\end{table*}

\subsection{Case Studies}

\begin{figure*}[h]
    \centering
        \includegraphics[width=0.96\linewidth]{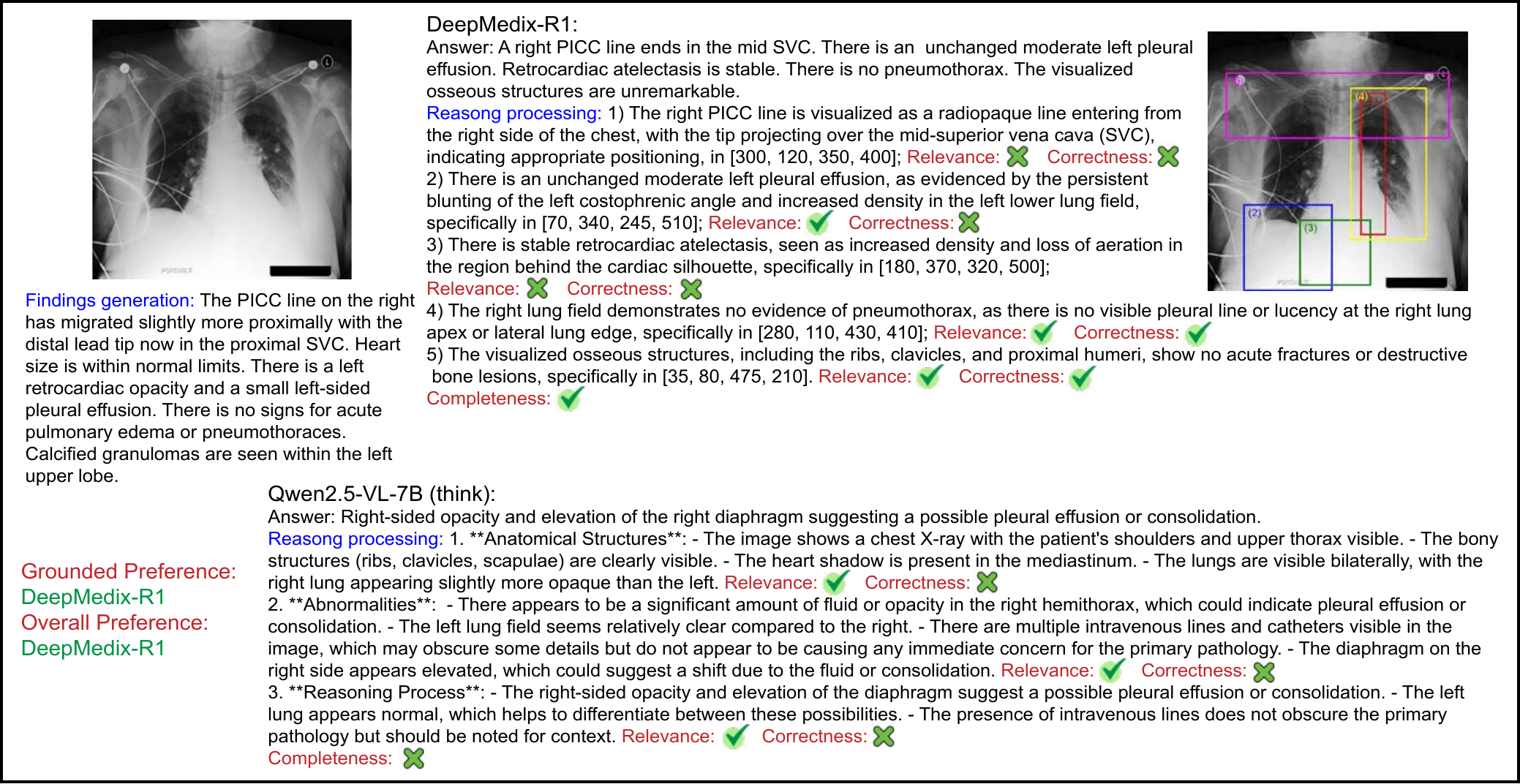}
    \caption{
    A case study for reasoning processing annotation across five dimensions, i.e., \emph{Relevance}, \emph{Correctness}, \emph{Completeness}, \emph{Grounded Preference}, and \emph{Overall Preference}.
    It is observed that \oursreasoning demonstrates superior performance in both grounded preference and overall preference for the reasoning process.
    }
    \label{fig_case1}
\end{figure*}

\begin{figure*}[h]
    \centering
        \includegraphics[width=0.96\linewidth]{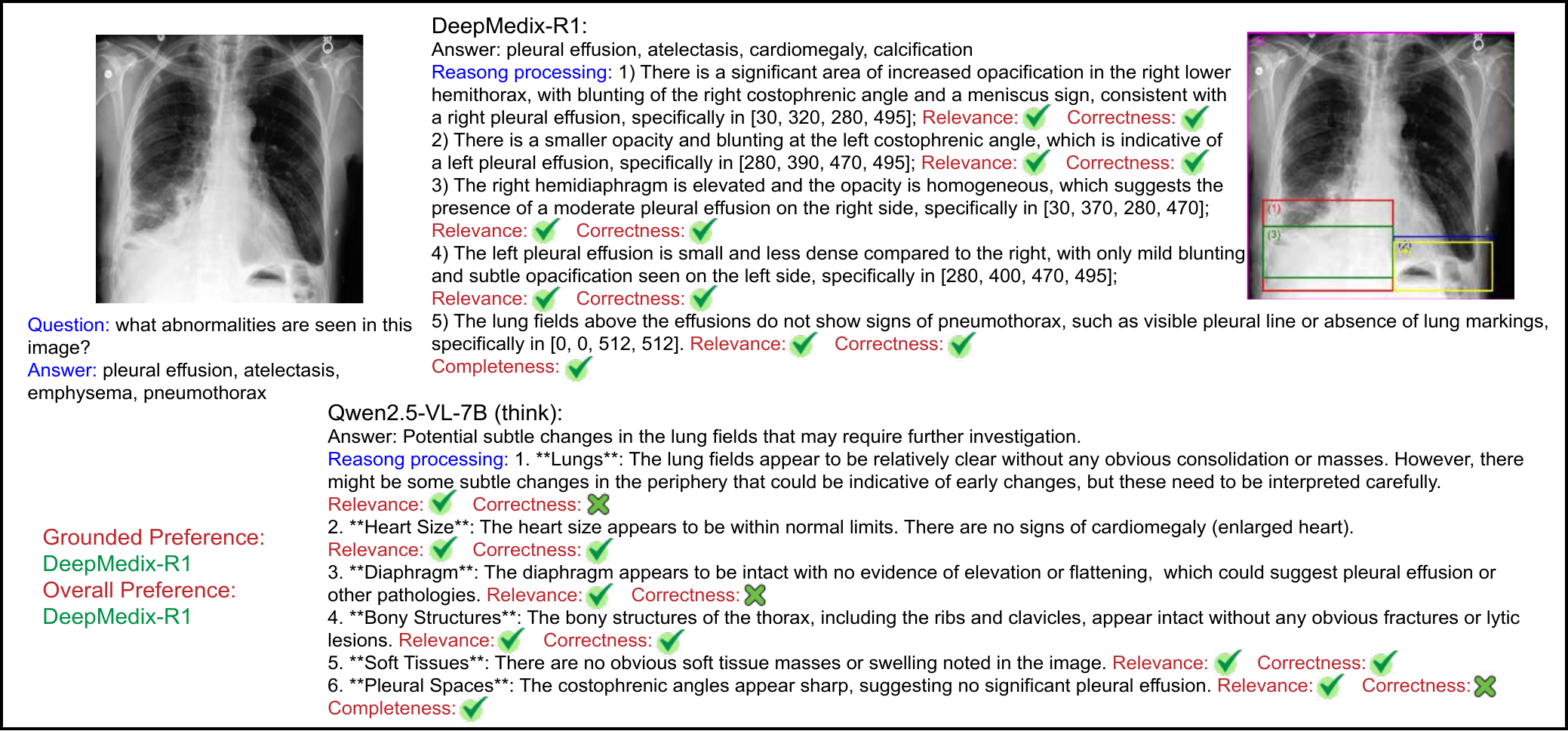}
    \caption{
    The second case study for reasoning processing annotation across five dimensions.
    }
    \label{fig_case2}
\end{figure*}

We present two representative case studies in Fig.~\ref{fig_case1} and \ref{fig_case2}, focusing on findings generation and VQA, respectively. These examples highlight the reasoning process across five key dimensions: Relevance, Correctness, Completeness, Grounded Preference, and Overall Preference. Additional case studies, provided in Tab.~\ref{tab_case_1}-\ref{tab_case_cxr14}, offer a comparative analysis of predictions from \oursreasoning, \ours, \huatuo, \chexagent, \llavarad, and \medgemma.

\section{Discussion}
\label{sec_discussion}

To enhance the performance, reasoning ability, and grounded perception of FMs for CXRs, we introduce the \oursreasoning model by instruction fine-tuning, cold start for reasoning, and group relative policy optimization stages. A comprehensive evaluation demonstrates that \oursreasoning outperforms state-of-the-art open-source FMs in both general and medical-specific domains.
Furthermore, online RL significantly improves both model performance and reasoning capabilities.

Based on the results, we propose four prospective directions to further advance the current medical FMs.
Firstly, as online RL is particularly beneficial for medical FMs, it enables them to continuously adapt and improve their performance based on real-time interactions and feedback.
As there is a wealth of domain knowledge in the medical field~\cite{zhang2023knowledge,yang2024poisoning}, it can be effectively leveraged to design a well-informed reward function for RL. By incorporating expert insights, structured rules, or heuristic-based metrics into the reward mechanism, the RL agent can receive more meaningful and interpretable feedback, guiding its learning process toward desired behaviors. This approach not only accelerates convergence by reducing sparse reward challenges but also enhances the model's alignment with domain-specific objectives, ensuring that the learned policies are both practical and high-performing in real-world applications.

Secondly, it can be observed from Tab.~\ref{tab_ext_data} that the FMs exhibit a high-recall but low-precision profile in external diagnostic prediction, e.g., 0.8748 and 0.6156 of \oursreasoning, and 0.9947 and 0.5012 of \llavarad.
This suggests that the FMs tend to respond ``yes'' more frequently to disease-related questions, thereby reducing the clinical risk of false negatives (missing true disease cases) which are generally more detrimental than false positives, where healthy individuals are incorrectly flagged.
However, such low-precision behavior limits the practical applicability of these models, as it undermines their clinical reliability and credibility.
This trade-off between recall and precision remains a common and challenging issue in FMs~\cite{le2024exploring}. Future research should aim to improve this balance by designing more advanced model architectures, fine-tuning on larger and better-curated medical datasets, and developing more effective methods for calibrating model confidence scores.

Thirdly, compared to general-purpose FMs designed for broad tasks that have achieved remarkable success, medical FMs tailored for healthcare applications still exhibit suboptimal performance in clinical settings~\cite{hu2024omnimedvqa,jin2024fairmedfm}.
To address this limitation, strategies such as building larger and more finely curated datasets, diversifying task-specific training objectives, and aggregating biased data from multiple institutions and hospitals can help mitigate performance gaps. By incorporating more comprehensive and representative data, along with targeted task design, medical FMs can achieve improved accuracy, robustness, and generalizability in real-world clinical scenarios. It can be seen that our model indeed enhances the reasoning capability by utilizing the LLM-synthesized reasoning data and online RL. However, it still suffers from the limitation of data quality.
Thus, while acquiring latent medical reasoning process data is highly resource-intensive, it remains indispensable for training explainable FMs.
A unified standard is also needed to make FMs easy to learn.

Finally, the annotation results indicate that the model achieves a \emph{Correctness} score of 0.7503 and a \emph{Completeness} score of 0.8592. While these metrics are significantly higher than baseline \qwenthink, they remain insufficient for real-world deployment due to the persistent issue of hallucination in current FMs~\cite{yang2025mitigating,heiman2025factchexcker,jiidom}. Even minor inaccuracies or omissions would lead to critical errors in high-stakes applications in medical and healthcare domains.
To mitigate this limitation, a substantial number of samples for the grounded understanding task can be incorporated during the fine-tuning phase.
Also, in the future, further advancements in training methodologies, such as RL with human feedback, adversarial validation, or hybrid neuro-symbolic approaches, will be essential to mitigate hallucinations and push correctness and completeness closer to the near-perfect levels required for reliable and scalable AI solutions.

\clearpage

\onecolumn

\clearpage
{\small
\begin{longtable}{c}
\caption{Case studies in \mimicfindings.}\label{tab_case_1}\\
\toprule
\begin{minipage}{0.15\textwidth}
\centering
\includegraphics[width=\linewidth]{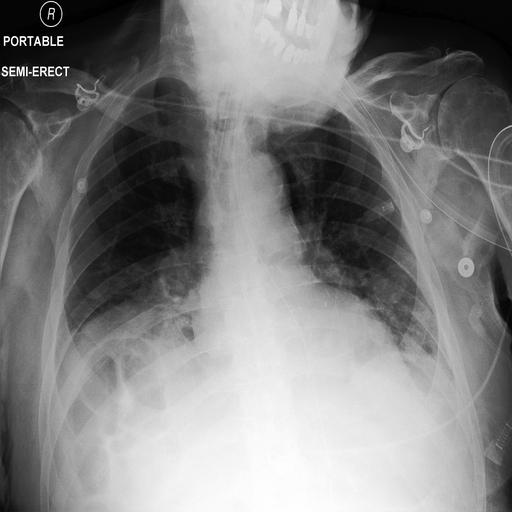}
\end{minipage}
\begin{minipage}{0.84\textwidth}
\textbf{Ground Truth:} A frontal semi-upright view of the chest was obtained portably. The endotracheal tube ends at the level of the inferior clavicular heads and is no less than 5.2cm above the carina. The nasogastric tube follows the expected course, although the tip is not visualized. Low lung volumes result in bronchovascular crowding. New opacification of both lower lobes despite no change in lung volumes compared to the prior study is concerning for pneuomonia. The upper lung zones are clear. There is no large pleural effusion or pneumothorax, although the left lung apex is obscured by the chin. The azygous vein is bigger than before with increased caliber of the left upper lobe vessels. The right hilus is chronically enlarged. Cardiac silhouette is stable. Flattening of the right humeral head may be due to avascular necrosis. Degenerative change is seen in the left shoulder girdle.
\end{minipage}
\\
\begin{minipage}{\textwidth}\textbf{\oursreasoning:} <think> 1) The endotracheal tube is visualized as a radiopaque line within the trachea, with its tip located just above the carina, confirming correct placement, specifically in \blue{[220, 70, 260, 420]}.
2) A nasogastric tube is seen coursing down the midline and passing below the diaphragm, with the tip out of the field of view, indicating correct positioning beyond the gastroesophageal junction, specifically in \blue{[260, 100, 320, 510]}.
3) There is increased density and blunting of both costophrenic angles, with layering along the lateral chest walls, indicating bilateral pleural effusions, specifically in \blue{[10, 350, 240, 510]} and, specifically in \blue{[270, 350, 490, 510]}.
4) Bibasilar opacities are visible, particularly at the lung bases bilaterally, raising suspicion for atelectasis, specifically in \blue{[100, 400, 400, 510]}.
5) The cardiomediastinal silhouette is within normal limits, showing no evidence of cardiomegaly or mediastinal widening, specifically in \blue{[180, 160, 340, 380]}. </think>. The answer is: \textbackslash boxed\{Single portable view of the chest. Endotracheal tube is now seen with tip approximately 2.5 cm from the carina. A nasogastric tube passes below the diaphragm with tip in the region of the gastric fundus. There are bilateral pleural effusions, larger on the right than on the left, with bibasilar opacities. The cardiomediastinal silhouette is unchanged.\}\\
\textbf{\ours:} As compared to the previous radiograph, there is no relevant change. The monitoring and support devices are constant. Constant appearance of the cardiac silhouette. Minimal atelectasis at the right lung bases. No evidence of pneumonia or pulmonary edema.
\\
\textbf{\huatuo:} The chest X-ray reveals several notable findings. The heart appears enlarged, suggesting cardiomegaly. There is evidence of bilateral pleural effusion, with fluid accumulation visible in both pleural spaces. Additionally, there are signs of pulmonary edema, characterized by increased interstitial markings and possible alveolar opacities. The lung fields show patchy areas of increased opacity, which could indicate consolidation or atelectasis. The diaphragm is elevated, and the costophrenic angles are blunted, further supporting the presence of pleural effusion. The mediastinum appears widened, which may be due to the enlarged heart or other mediastinal pathology. Overall, these findings suggest significant cardiac and pulmonary abnormalities that warrant further clinical correlation and investigation.
\\
\textbf{\chexagent:} Left lower lung zone, left lung, right lower lung zone, right lung
\\
\textbf{\llavarad:} The patient is intubated. The endotracheal tube terminates about 3.5 cm above the carina. An orogastric tube courses into the stomach. The cardiac, mediastinal and hilar contours appear unchanged. There is persistent opacification at the medial right lung base. A patchy new left basilar opacity is also noted. There is no definite pleural effusion or pneumothorax.
\\
\textbf{\medgemma:} **FINDINGS:**
The patient is in a portable semi-erect position. The image is limited by the patient's positioning and the presence of an endotracheal tube.
*   **Lungs:** There is increased opacity in the lower lung fields bilaterally, which could be due to atelectasis, consolidation, or pleural effusion. The upper lung fields appear relatively clear.
*   **Heart:** The heart size is difficult to assess due to the patient's positioning and the presence of the endotracheal tube.
*   **Mediastinum:** The mediastinum appears unremarkable.
*   **Bones:** The visualized bony structures are unremarkable.
*   **Endotracheal Tube:** An endotracheal tube is present, terminating approximately 4 cm above the carina.
**Overall Impression:**
The image is limited, and further evaluation with a lateral radiograph and/or CT scan is recommended to better assess the lung fields and rule out any significant pathology.
**Disclaimer:** This is a simulated report based on the provided image. A real radiologist would need to review the full clinical context and potentially additional imaging to provide a definitive diagnosis.
\end{minipage}\\
\midrule
\begin{minipage}{0.15\textwidth}
\centering
\includegraphics[width=\linewidth]{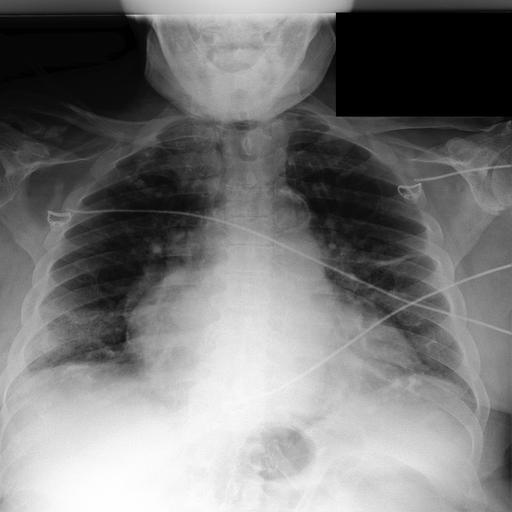}
\end{minipage}
\begin{minipage}{0.84\textwidth}
\textbf{Ground Truth:} The heart is moderately enlarged. The aortic arch is calcified. The mediastinal and hilar contours appear unchanged. The lung volumes are low. Calcified pleural plaques are present. There is no definite pleural effusion or pneumothorax. Band-like opacity in the left mid lung suggests minor atelectasis or scarring. Pulmonary vessels are somewhat engorged centrally suggesting pulmonary venous hypertension if not frank pulmonary edema. There is a confluent right basilar opacity worrisome for pneumonia.
\end{minipage}\\
\makecell[{{p{\textwidth}}}]{\textbf{\oursreasoning:} <think> 1) The cardiac silhouette is clearly enlarged, occupying more than half of the thoracic width, indicating moderate cardiomegaly, specifically in \blue{[160, 210, 340, 450]}.
2) The pulmonary vasculature appears prominent, particularly in the hilar and upper lung zones, with increased vascular markings, consistent with mild pulmonary vascular congestion, specifically in \blue{[80, 100, 400, 350]}.
3) There are linear opacities at the lung bases bilaterally, which are suggestive of bibasilar atelectasis, specifically in \blue{[60, 400, 210, 500]} and, specifically in \blue{[300, 400, 470, 500]}.
4) The costophrenic angles are not sharply visualized, and there is blunting and increased density at the lung bases, indicating probable small bilateral pleural effusions, specifically in \blue{[80, 430, 170, 500]} and, specifically in \blue{[340, 430, 440, 500]}.
5) The lung fields demonstrate no focal areas of increased opacity or consolidation, ruling out focal consolidation, specifically in \blue{[80, 100, 440, 400]}.
6) No evidence of a radiolucent line or absence of lung markings is seen at the lung periphery, indicating there is no evidence of pneumothorax, specifically in \blue{[40, 100, 480, 500]}. </think>. The answer is: \textbackslash boxed\{Single AP portable view of the chest was obtained. There is moderate cardiomegaly. There is mild pulmonary vascular congestion. Bibasilar atelectasis is seen. Small bilateral pleural effusions are likely present. No focal consolidation or pneumothorax.\}}
\\
\makecell[{{p{\textwidth}}}]{\textbf{\ours:} Single portable view of the chest. There is moderate cardiomegaly. There is no evidence of pulmonary edema. There is no pleural effusion or pneumothorax. No definite focal consolidation is seen.}
\\
\makecell[{{p{\textwidth}}}]{\textbf{\huatuo:} The chest X-ray reveals several notable findings. The heart appears enlarged, suggesting cardiomegaly. There is evidence of bilateral pleural effusion, with fluid accumulation visible in both pleural cavities. Additionally, there is a moderate amount of ascites, indicated by the presence of free fluid in the abdominal cavity. The lung fields show signs of pulmonary congestion, which may be due to underlying cardiac or renal conditions. The diaphragm and costophrenic angles are obscured, further supporting the presence of pleural effusions. Overall, these findings suggest significant cardiovascular and potentially hepatic involvement.}
\\
\makecell[{{p{\textwidth}}}]{\textbf{\chexagent:} Enlarged cardiac silhouette, lung opacity, pulmonary edema/hazy opacity, vascular congestion}
\\
\makecell[{{p{\textwidth}}}]{\textbf{\llavarad:} The lung volumes are low. The heart is moderately enlarged. The aortic arch is calcified. The mediastinal and hilar contours are probably unchanged. There is no pleural effusion or pneumothorax. A new opacity in the right lower lung raises concern for pneumonia. Streaky left mid lung opacities are also noted and could be seen with atelectasis.}
\\
\makecell[{{p{\textwidth}}}]{\textbf{\medgemma:} The image shows a chest X-ray. The heart appears enlarged, with increased prominence of the cardiac silhouette. The mediastinum is also somewhat prominent. The lung fields are relatively clear, with no obvious large consolidations or effusions. There is a possible linear opacity in the left mid-lung field, which could represent a small area of atelectasis or scarring. The visualized bony structures appear intact.
**In summary:**
*   **Cardiomegaly:** The heart appears enlarged.
*   **Mediastinal prominence:** The mediastinum is somewhat prominent.
*   **Possible atelectasis/scarring:** A linear opacity is present in the left mid-lung field.
**Important Considerations:**
*   This is a limited assessment based on a single image.
*   Further evaluation, including comparison to prior imaging and clinical correlation, is necessary for a definitive diagnosis.
*   The presence of a linear opacity in the left mid-lung field warrants further investigation to rule out potential underlying pathology. 
**Disclaimer:** This report is for informational purposes only and should not be considered a substitute for a professional medical diagnosis. A qualified healthcare professional should interpret the image and provide a complete medical evaluation.}
\\
\midrule
\begin{minipage}{0.15\textwidth}
\centering
\includegraphics[width=\linewidth]{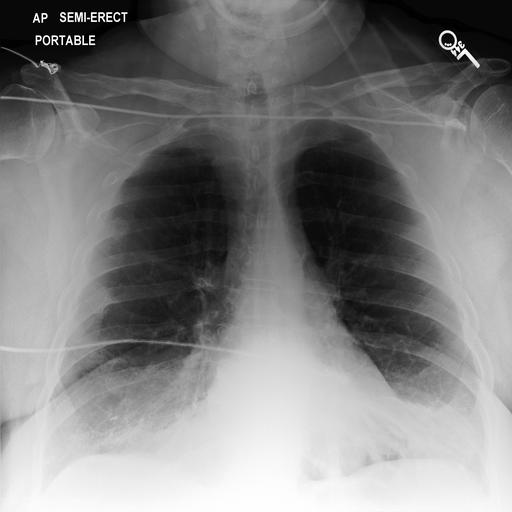}
\end{minipage}
\begin{minipage}{0.84\textwidth}
\textbf{Ground Truth:} Heart size is normal. Mediastinal and hilar contours are unremarkable. The pulmonary vascularity is normal. Streaky bibasilar airspace opacities likely reflect atelectasis. No pleural effusion or pneumothorax is seen. Multiple old right-sided rib fractures are re-visualized.
\end{minipage}\\
\makecell[{{p{\textwidth}}}]{\textbf{\oursreasoning:} <think> 1) The cardiac silhouette and mediastinal contours are visualized in the central region of the image, specifically in \blue{[180, 100, 330, 400]}. These appear within normal limits, with no evidence of cardiomegaly or abnormal widening.
2) Examination of the lung fields bilaterally, specifically in \blue{[50, 100, 200, 450]} (left lung) and, specifically in \blue{[330, 100, 470, 450]} (right lung) reveals no focal areas of consolidation, as the lung parenchyma appears clear without focal opacities.
3) The pleural spaces at the lung apices and costophrenic angles, specifically in \blue{[60, 400, 210, 500]} (left) and, specifically in \blue{[330, 400, 470, 500]} (right) show no evidence of pneumothorax (absence of a visible pleural line) or pleural effusion (no blunting or meniscus sign).
4) The bony thorax, including the ribs, clavicles, and visible parts of the spine, specifically in \blue{[30, 50, 480, 500]}, appears intact without any fractures or destructive lesions. </think>. The answer is: \textbackslash boxed\{Single AP semi-upright portable view of the chest was obtained. The cardiac and mediastinal silhouettes are unremarkable. No focal consolidation, pleural effusion, or evidence of pneumothorax is seen. No overt pulmonary edema is seen. There is no evidence of free air beneath the diaphragms.\}}
\\
\makecell[{{p{\textwidth}}}]{\textbf{\ours:} As compared to the previous radiograph, there is no relevant change. The lung volumes remain normal. There is no evidence of pneumonia or other acute lung disease. No pleural effusions. No pulmonary edema. Normal size of the cardiac silhouette.}
\\
\makecell[{{p{\textwidth}}}]{\textbf{\huatuo:} The chest X-ray shows clear lung fields with no evidence of consolidation, pleural effusion, or pneumothorax. The heart size is within normal limits, and the mediastinal contours appear normal. There are no visible masses or nodules. The diaphragm and costophrenic angles are sharp and well-defined. The bony thoracic cage, including the ribs and clavicles, appears intact without any fractures or lesions. Overall, the radiographic findings suggest a normal chest X-ray with no acute abnormalities detected.}
\\
\makecell[{{p{\textwidth}}}]{\textbf{\chexagent:} Right lower lung zone, right lung, right mid lung zone.}
\\
\makecell[{{p{\textwidth}}}]{\textbf{\llavarad:} Single portable view of the chest. Exam is limited due to portable technique and patient's arms overlying the chest. The lungs are grossly clear. The cardiomediastinal silhouette is within normal limits. Old healed right lateral rib fractures are identified.}
\\
\makecell[{{p{\textwidth}}}]{\textbf{\medgemma:} Okay, I will act as an experienced radiologist and provide a "FINDINGS" section based on the provided AP semi-erect portable chest X-ray.
**FINDINGS:**
The lungs are clear bilaterally without evidence of focal consolidation, pleural effusion, or pneumothorax. The heart size is within normal limits. The mediastinal contours are unremarkable. The bony structures of the thorax are intact. There is no evidence of acute trauma.
**Impression:**
No acute cardiopulmonary abnormalities identified.}
\\
\bottomrule
\end{longtable}
}

\clearpage
{\small
\begin{longtable}{c}
\caption{Case studies in \mimicimpression.}\label{tab_case_2}\\
\toprule
\begin{minipage}{0.15\textwidth}
\centering
\includegraphics[width=\linewidth]{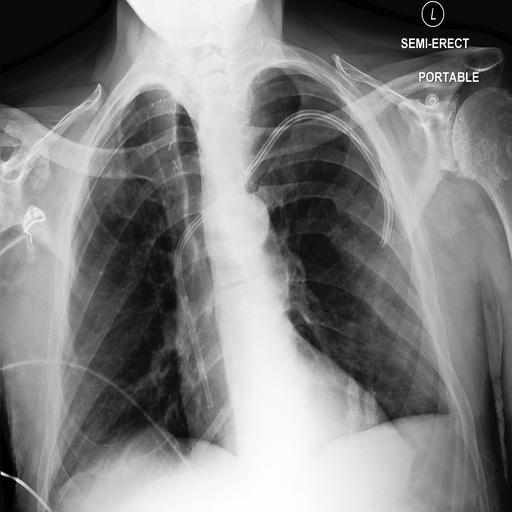}
\end{minipage}
\begin{minipage}{0.84\textwidth}
\textbf{Ground Truth:} No definite evidence to suggest pneumonia or fluid overload.
\end{minipage}\\
\makecell[{{p{\textwidth}}}]{\textbf{\oursreasoning:} <think> 1) The left-sided central venous catheter and right central line are visualized in appropriate positions, with their tips projecting over the expected locations of the superior vena cava and right atrium, respectively, specifically in \blue{[120, 70, 400, 400]}.
2) The cardiac silhouette appears normal in size and contour, with no evidence of cardiomegaly or abnormal cardiac borders, specifically in \blue{[160, 250, 350, 430]}.
3) The bilateral lung fields are clear, with no focal consolidation, mass, or interstitial opacities, suggesting there is no pneumonia or pulmonary edema, specifically in \blue{[60, 120, 440, 480]}.
4) The costophrenic angles are sharp and free of blunting or fluid, indicating no pleural effusion, specifically in \blue{[70, 420, 420, 500]}.
5) The mediastinal contours are normal, with no widening or abnormal masses, specifically in \blue{[180, 100, 320, 320]}. </think>. The answer is: \textbackslash boxed\{As compared to the previous radiograph, the left-sided central venous access line and the right internal jugular vein catheter are in unchanged position. Normal size of the cardiac silhouette. No pleural effusions. No pneumonia, no pulmonary edema.\}}
\\
\makecell[{{p{\textwidth}}}]{\textbf{\ours:} No acute cardiopulmonary process.}
\\
\makecell[{{p{\textwidth}}}]{\textbf{\huatuo:} The chest X-ray reveals a right-sided pneumothorax with a significant volume of air in the pleural space, causing partial collapse of the right lung. There is also evidence of a right-sided hemothorax, indicated by the presence of fluid in the pleural cavity. The mediastinum appears to be shifted slightly towards the left side, suggesting tension pneumothorax. These findings are consistent with a traumatic injury, likely due to a gunshot wound. The patient's clinical history of a gunshot wound to the right chest wall supports these radiographic observations.}
\\
\makecell[{{p{\textwidth}}}]{\textbf{\chexagent:} The left chest tube has been removed. There is no evidence of pneumothorax. The lungs are clear. The cardiomediastinal silhouette is within normal limits. Osseous and soft tissue structures are unremarkable.}
\\
\makecell[{{p{\textwidth}}}]{\textbf{\llavarad:} A left-sided dialysis catheter terminates in the right atrium. The heart size is normal. The hilar and mediastinal contours are within normal limits. There is no pneumothorax, focal consolidation, or pleural effusion. Multiple old left-sided rib fractures are present.}
\\
\makecell[{{p{\textwidth}}}]{\textbf{\medgemma:} **IMPRESSION:**
*   **Central Venous Catheters:** Two central venous catheters are present, one in the superior vena cava and one in the right atrium.
*   **Mediastinal Mass:** There is a prominent mediastinal mass, likely representing a lymph node or other mediastinal structure.
*   **No Obvious Pneumothorax:** No large pneumothorax is identified.
*   **No Obvious Pleural Effusion:** No large pleural effusion is identified.
*   **No Obvious Lung Consolidation:** No obvious lung consolidation is identified.
**Potential Diagnoses:**
*   Mediastinal mass (lymphadenopathy, tumor, etc.)
*   Possible mild pulmonary vascular congestion.}
\\
\midrule
\begin{minipage}{0.15\textwidth}
\centering
\includegraphics[width=\linewidth]{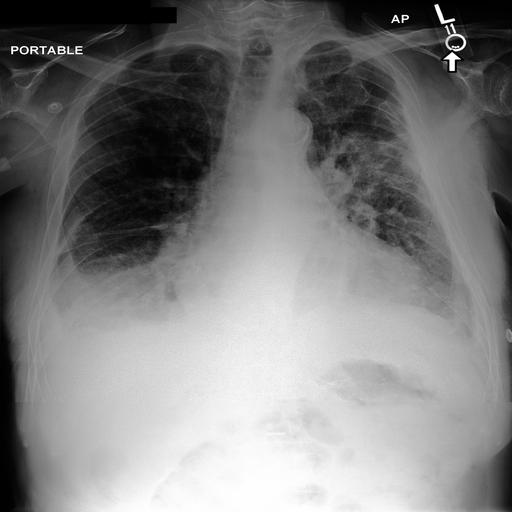}
\end{minipage}
\begin{minipage}{0.84\textwidth}
\textbf{Ground Truth:} New small bilateral pleural effusions with mild-to-moderate pulmonary edema.
\end{minipage}\\
\makecell[{{p{\textwidth}}}]{\textbf{\oursreasoning:} <think> 1) The left lung field, particularly in the mid and lower zones, shows increased hazy opacities and interstitial markings, which are indicative of pulmonary edema, specifically in \blue{[100, 150, 450, 450]};
2) The right lung base, specifically in the lower right region, demonstrates a more coalescent opacity compared to the left, raising suspicion for a developing consolidation which could represent pneumonia, specifically in \blue{[400, 400, 500, 500]};
3) There is blunting of the right costophrenic angle with increased density at the right lung base, consistent with a small right pleural effusion, specifically in \blue{[400, 450, 500, 500]};
4) The heart size appears mildly enlarged, which is suggestive of mild cardiomegaly, specifically in \blue{[200, 300, 350, 450]}. </think>. The answer is: \textbackslash boxed\{As compared to the previous image, the left lung shows an increase in severity of the pre-existing pulmonary edema. The right lung base shows a small pleural effusion, combined to a more coalescent opacity, potentially reflecting pneumonia. Mild cardiomegaly persists.\}}
\\
\makecell[{{p{\textwidth}}}]{\textbf{\ours:} No significant interval change when compared to the prior study.}
\\
\makecell[{{p{\textwidth}}}]{\textbf{\huatuo:} The chest X-ray reveals bilateral pulmonary infiltrates with a predominance in the right lung. There is evidence of pneumomediastinum, indicated by the presence of air in the mediastinal region. Additionally, there is a small amount of free air under the diaphragm, suggesting possible gastrointestinal perforation or other causes of pneumoperitoneum. These findings suggest a serious underlying condition that may require immediate medical attention.}
\\
\makecell[{{p{\textwidth}}}]{\textbf{\chexagent:} Unchanged appearance of the lung parenchyma. Unchanged appearance of the cardiac silhouette.}
\\
\makecell[{{p{\textwidth}}}]{\textbf{\llavarad:} A right-sided PICC line terminates in the mid superior vena cava. The heart is mildly enlarged. The aortic arch is calcified. There is a moderate pleural effusion on the right and a small one on the left. Patchy opacities in the left upper lobe and in the right lower lung are probably compatible with atelectasis. There is no pneumothorax.}
\\
\makecell[{{p{\textwidth}}}]{\textbf{\medgemma:} Right pleural effusion, likely with associated atelectasis. Left lung shows increased interstitial markings and possible consolidation in the upper lobe, concerning for pneumonia or other inflammatory process. Cardiac silhouette is enlarged. Right-sided central line is present.
}
\\
\midrule
\begin{minipage}{0.15\textwidth}
\centering
\includegraphics[width=\linewidth]{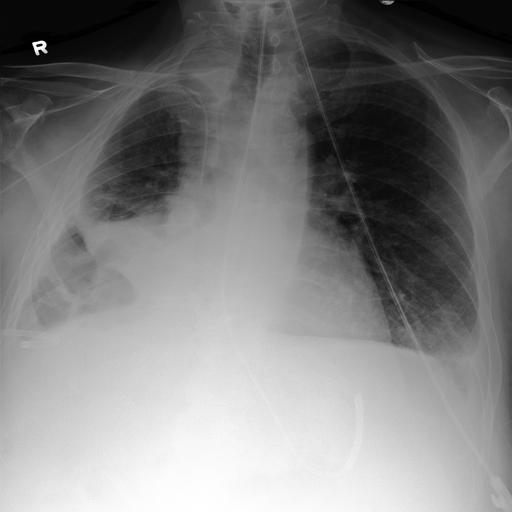}
\end{minipage}
\begin{minipage}{0.84\textwidth}
\textbf{Ground Truth:} 1. Dobbhoff tube in the stomach. 2. Unchanged right basilar loculated hydropneumothorax.
\end{minipage}\\
\makecell[{{p{\textwidth}}}]{\textbf{\oursreasoning:} <think> 1) The course of the Dobbhoff tube can be traced from the upper chest down through the midline toward the stomach region, confirming its position within the stomach, specifically in \blue{[230, 100, 315, 480]};
2) The right internal jugular line and right subclavian line are visualized coursing from the neck, with their tips appropriately located in the superior vena cava and right atrium, respectively, specifically in \blue{[35, 60, 160, 230]};
3) There is a large right pleural effusion, demonstrated by the homogenous opacity occupying much of the right lower hemithorax and blunting of the right costophrenic angle, specifically in \blue{[30, 230, 250, 470]};
4) A small left pleural effusion is evident as blunting of the left costophrenic angle and increased density in the left lower lung field, specifically in \blue{[270, 400, 480, 500]};
5) There is a large mass in the right upper lobe, which appears as a dense, irregular opacity in the right upper lung region, specifically in \blue{[110, 120, 220, 250]}. </think>. The answer is: \textbackslash boxed\{As compared to the previous radiograph, the Dobbhoff tube is now in correct position in the stomach. The right internal jugular line and the right subclavian line are also in correct position. The large right pleural effusion and the small left pleural effusion are unchanged. Unchanged large right upper lobe mass.\}}
\\
\makecell[{{p{\textwidth}}}]{\textbf{\ours:} Dobbhoff tube terminates in the stomach.}
\\
\makecell[{{p{\textwidth}}}]{\textbf{\huatuo:} The chest X-ray reveals a large, homogenous opacity in the right hemithorax, consistent with a massive pleural effusion. The heart is shifted to the left, indicating mediastinal shift due to the effusion. There is no evidence of lung collapse or consolidation within the effusion. The left lung appears normal without any signs of pathology. These findings suggest a significant accumulation of fluid in the pleural space on the right side, likely due to an underlying condition such as infection, malignancy, or heart failure. Further evaluation with ultrasound or CT scan may be necessary for definitive diagnosis and management planning.}
\\
\makecell[{{p{\textwidth}}}]{\textbf{\chexagent:} In comparison with the earlier study of this date, there has been placement of a Dobbhoff tube that extends to the lower body of the stomach. Otherwise, little change.}
\\
\makecell[{{p{\textwidth}}}]{\textbf{\llavarad:} The Dobbhoff tube is coiled in the stomach. The right PICC line is unchanged. The right pigtail catheter is unchanged. There is no pneumothorax. There is a small left pleural effusion. There is a stable right loculated pleural effusion. The right lower lobe atelectasis is unchanged.}
\\
\makecell[{{p{\textwidth}}}]{\textbf{\medgemma:} Right pleural effusion, likely loculated, with associated atelectasis. Left lung appears relatively clear. Right-sided picc line is present. Consider thoracentesis for diagnostic evaluation of the right pleural effusion.}
\\
\bottomrule
\end{longtable}
}

\clearpage
{\small
\begin{longtable}{c}
\caption{Case studies in \openifindings.}\label{tab_case_3}\\
\toprule
\begin{minipage}{0.15\textwidth}
\centering
\includegraphics[width=\linewidth]{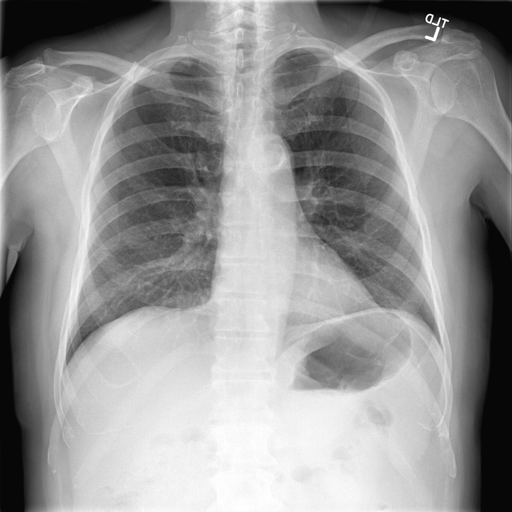}
\end{minipage}
\begin{minipage}{0.84\textwidth}
\textbf{Ground Truth:} The lungs are clear. No pleural effusion is identified. The heart is normal. There are calcifications of the aortic XXXX. The skeletal structures are normal.
\end{minipage}\\
\makecell[{{p{\textwidth}}}]{\textbf{\oursreasoning:} <think> 1) The cardiomediastinal silhouette and pulmonary vasculature appear in the central region of the image, showing no evidence of acute enlargement, displacement, or abnormal contour. The heart size is within normal limits, specifically in \blue{[180, 110, 340, 370]}
2) The lungs are examined for focal areas of increased opacity or airspace disease (such as consolidation), and none are seen. No visible nodular or mass-like densities are identified, specifically in \blue{[60, 120, 210, 400]} and, specifically in \blue{[340, 120, 480, 400]}
3) Examination of the pleural spaces reveals clear costophrenic angles without evidence of pleural effusion or pneumothorax. No visible pleural thickening or free air is identified, specifically in \blue{[60, 390, 180, 480]} and, specifically in \blue{[340, 390, 480, 480]}
4) The costophrenic angles and diaphragms are visualized and appear sharp and well-defined, suggesting no acute diaphragmatic or pleural abnormalities, specifically in \blue{[70, 400, 190, 490]} and, specifically in \blue{[340, 400, 470, 490]}
5) The bony thorax, including ribs, clavicles, clavicular heads, and spine, shows no evidence of acute fracture, destructive lesion, or other abnormality, specifically in \blue{[30, 30, 480, 490]} </think>. The answer is: \textbackslash boxed\{The cardiomediastinal silhouette and pulmonary vasculature are within normal limits in size and contour. The lungs are clear of focal airspace disease, pneumothorax, or pleural effusion. There are no acute bony findings.\}}
\\
\makecell[{{p{\textwidth}}}]{\textbf{\ours:} The lungs are clear. There is no pleural effusion or pneumothorax. The heart and mediastinum are normal. The skeletal structures are normal.}
\\
\makecell[{{p{\textwidth}}}]{\textbf{\huatuo:} **FINDINGS:**
1. **Lung Fields:** The lung fields appear clear without evidence of consolidation, masses, or significant pleural effusions. The lung parenchyma shows normal vascular markings with no signs of interstitial changes.
2. **Heart and Mediastinum:** The cardiac silhouette is within normal limits for size and contour, suggesting no cardiomegaly. The mediastinal structures appear unremarkable, with no widening or abnormal masses noted.
3. **Diaphragm and Costophrenic Angles:** The diaphragms are well-defined and positioned normally. The costophrenic angles are sharp, indicating no pleural effusion.
4. **Bones:** The bony thorax, including ribs, clavicles, and vertebrae, appears intact without fractures or lesions. The shoulder joints show normal alignment.
5. **Soft Tissues:** There are no visible abnormalities in the soft tissues surrounding the thoracic cavity.
6. **Pleural Spaces:** No pleural effusions are evident, and the pleural lines are smooth and continuous.
7. **Overall Assessment:** The chest X-ray demonstrates normal findings with no acute pathological changes. The patient's lungs, heart, and mediastinal structures appear to be within normal limits.
**Conclusion:** The chest X-ray is unremarkable, showing no significant abnormalities.}
\\
\makecell[{{p{\textwidth}}}]{\textbf{\chexagent:} Left lower lung zone, left lung, left mid lung zone, left upper lung zone, right lower lung zone, right lung, right mid lung zone, right upper lung zone}
\\
\makecell[{{p{\textwidth}}}]{\textbf{\llavarad:} The heart is normal in size and there is no vascular congestion, pleural effusion, or acute focal pneumonia.}
\\
\makecell[{{p{\textwidth}}}]{\textbf{\medgemma:} **FINDINGS:**
*   **Heart Size:** The heart size is within normal limits. The cardiothoracic ratio appears to be less than 0.5.
*   **Mediastinum:** The mediastinum is unremarkable in appearance.
*   **Lungs:** The lungs are clear bilaterally, without evidence of focal consolidation, pleural effusion, or pneumothorax. There is no obvious pulmonary edema.
*   **Bones:** The visualized bony structures (ribs, clavicles, scapulae, and humeri) appear intact. There are no obvious fractures.
*   **Abdomen:** The bowel loops are present and appear unremarkable.
*   **Other:** No other significant findings are apparent on this single frontal view.}
\\
\midrule
\begin{minipage}{0.15\textwidth}
\centering
\includegraphics[width=\linewidth]{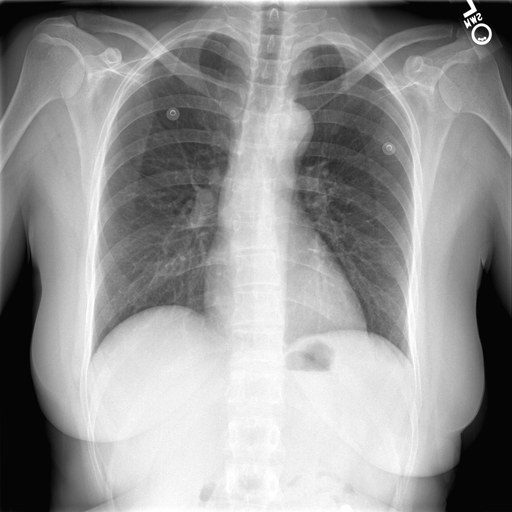}
\end{minipage}
\begin{minipage}{0.84\textwidth}
\textbf{Ground Truth:} The cardiomediastinal silhouette and pulmonary vasculature are within normal limits in size. The lungs are clear of focal airspace disease, pneumothorax, or pleural effusion. There are no acute bony findings.
\end{minipage}\\
\makecell[{{p{\textwidth}}}]{\textbf{\oursreasoning:} <think> 1) The cardiomediastinal silhouette and pulmonary vasculature are visualized in the central region of the image, appearing within normal limits in size and contour. There is no evidence of cardiomegaly or abnormal vascular markings, specifically in \blue{[180, 90, 320, 400]}
2) The lungs are clear, without areas of focal consolidation, mass, or interstitial markings suggestive of infiltrate. No pleural effusion or pneumothorax is present, specifically in \blue{[80, 100, 420, 400]}
3) The osseous structures, including the ribs, clavicles, scapulae, and visualized parts of the spine, are examined and show no acute fractures, destructive lesions, or deformities, specifically in \blue{[30, 30, 470, 480]}
4) The diaphragm and costophrenic XXXX are smooth and well-defined, with no evidence of blunting or free subdiaphragmatic air, specifically in \blue{[100, 380, 400, 490]} </think>. The answer is: \textbackslash boxed\{The cardiomediastinal silhouette and pulmonary vasculature are within normal limits in size and contour. The lungs are clear of focal airspace disease, pneumothorax, or pleural effusion. There are no acute bony findings.\}}
\\
\makecell[{{p{\textwidth}}}]{\textbf{\ours:} The lungs are clear. There is no pleural effusion or pneumothorax. The heart and mediastinum are normal. The skeletal structures are normal.}
\\
\makecell[{{p{\textwidth}}}]{\textbf{\huatuo:} **FINDINGS:**
1. **Pleural Effusion:** There is a noticeable opacity in the left hemithorax, indicative of pleural effusion. This fluid accumulation is causing a shift of the mediastinal structures towards the right side.
2. **Mediastinal Shift:** The mediastinum appears to be shifted towards the right, which is consistent with the presence of the pleural effusion on the left side.
3. **Lung Fields:** The lung fields appear relatively clear on the right side, but there is evidence of compression atelectasis or reduced volume on the left due to the effusion.
4. **Cardiac Silhouette:** The cardiac silhouette is visible and appears normal in size and contour, without any obvious signs of cardiomegaly.
5. **Diaphragm:** The diaphragm on the left side is obscured by the effusion, making it difficult to assess its contour. On the right side, the diaphragm appears normal.
6. **Bones and Soft Tissues:** The bony thoracic cage and soft tissues appear unremarkable without any visible fractures or abnormalities.
7. **Electrodes:** There are visible electrodes on both shoulders, likely for monitoring purposes during the imaging procedure.}
\\
\makecell[{{p{\textwidth}}}]{\textbf{\chexagent:} Left lower lung zone, left lung, left mid lung zone, left upper lung zone, right lower lung zone, right lung, right mid lung zone, right upper lung zone.}
\\
\makecell[{{p{\textwidth}}}]{\textbf{\llavarad:} The heart is normal in size and there is no vascular congestion, pleural effusion, or acute focal pneumonia.}
\\
\makecell[{{p{\textwidth}}}]{\textbf{\medgemma:} **FINDINGS:**
The lungs are clear bilaterally without evidence of focal consolidation, pleural effusion, or pneumothorax. The heart size is normal. The mediastinal contours are unremarkable. The bony structures of the chest wall are intact. No acute fractures are identified. The visualized upper abdomen is unremarkable.
**In summary, the chest X-ray demonstrates no acute abnormalities.**
**Important Considerations:**
*   This is a limited assessment based on a single frontal view. A lateral view would be helpful for a more comprehensive evaluation.
*   The image quality is adequate for the assessment provided.
*   This report is for informational purposes only and should not be used to make medical decisions. A qualified healthcare professional should interpret the X-ray in the context of the patient's clinical presentation and medical history.}
\\
\bottomrule
\end{longtable}
}

\clearpage
{\small
\begin{longtable}{c}
\caption{Case studies in \openiimporession.}\label{tab_case_4}\\
\toprule
\begin{minipage}{0.15\textwidth}
\centering
\includegraphics[width=\linewidth]{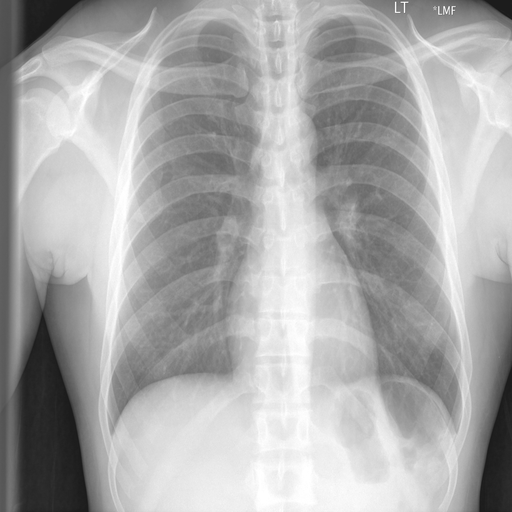}
\end{minipage}
\begin{minipage}{0.84\textwidth}
\textbf{Ground Truth:} Clear lungs. No effusions. Unremarkable mediastinal contour. No acute cardiopulmonary abnormality identified.
\end{minipage}\\
\makecell[{{p{\textwidth}}}]{\textbf{\oursreasoning:} <think> 1) The lung fields are evaluated for any areas of increased opacity, which could suggest consolidation or infiltrate. Both lungs appear clear, without focal consolidation or suspicious opacities, specifically in \blue{[100, 100, 400, 400]}
2) The cardiac silhouette is assessed for enlargement or abnormal contour. The heart size is within normal limits, and there are no signs of cardiomegaly or abnormal borders, specifically in \blue{[210, 260, 330, 420]}
3) The costophrenic angles and diaphragms are checked for blunting or abnormal elevation. Both costophrenic XXXX are sharp and the diaphragms are at normal levels, without evidence of pleural effusion or subdiaphragmatic abnormality, specifically in \blue{[120, 410, 390, 500]}
4) The bony thorax and soft tissues are reviewed for fractures, destructive lesions, or other abnormalities. The bones and soft tissues appear intact with no evidence of acute abnormality, specifically in \blue{[40, 40, 470, 500]}
Based on the above regional assessments, there is no acute cardiopulmonary abnormality. </think>. The answer is: \textbackslash boxed\{No acute cardiopulmonary abnormality.\}}
\\
\makecell[{{p{\textwidth}}}]{\textbf{\ours:} No acute cardiopulmonary abnormality.}
\\
\makecell[{{p{\textwidth}}}]{\textbf{\huatuo:} **IMPRESSION:**
1. **Pleural Effusion:** There is a moderate to large right-sided pleural effusion with associated atelectasis of the right lung. This is evidenced by the blunting of the right costophrenic angle and increased opacity in the right hemithorax.
2. **Cardiac Enlargement:** The cardiac silhouette appears enlarged, suggesting cardiomegaly. This could be due to underlying heart disease or fluid overload.
3. **Diaphragmatic Elevation:** The right diaphragm is elevated compared to the left, likely secondary to the pleural effusion and atelectasis.
4. **No Other Significant Abnormalities:** The left lung field appears clear, and there are no other obvious abnormalities such as masses, nodules, or significant pulmonary infiltrates.}
\\
\makecell[{{p{\textwidth}}}]{\textbf{\chexagent:} The Frontal view of the X-ray is unremarkable.}
\\
\makecell[{{p{\textwidth}}}]{\textbf{\llavarad:} The lungs are clear without infiltrate or effusion. The bony thorax is normal. Mediastinal and cardiac silhouettes are normal.}
\\
\makecell[{{p{\textwidth}}}]{\textbf{\medgemma:} The lungs are clear bilaterally. The heart size is normal. The mediastinum is unremarkable. No pleural effusion or pneumothorax is identified.}
\\
\midrule
\begin{minipage}{0.15\textwidth}
\centering
\includegraphics[width=\linewidth]{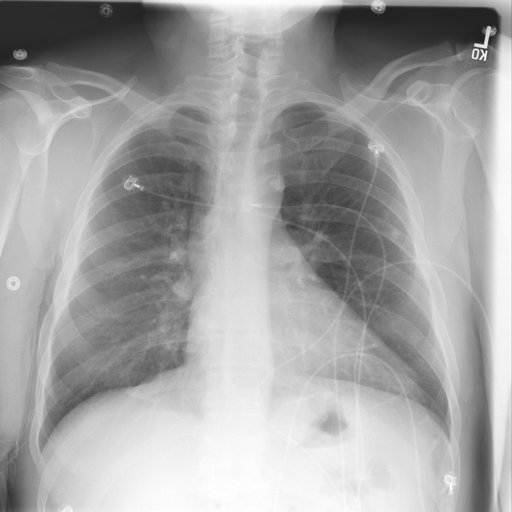}
\end{minipage}
\begin{minipage}{0.84\textwidth}
\textbf{Ground Truth:} Mediastinal cardiac size stable. No pneumothoraces or large pleural effusions. Pulmonary granulomas. No acute pulmonary process. No acute displaced rib fractures.
\end{minipage}\\
\makecell[{{p{\textwidth}}}]{\textbf{\oursreasoning:} <think> 1) The cardiac silhouette and mediastinal contours appear within normal limits in size and shape, with no evidence of enlargement or abnormal contour, suggesting no acute cardiopulmonary abnormality, specifically in \blue{[180, 150, 335, 410]}
2) The lung fields are clear, with no focal areas of consolidation, interstitial markings, or suspicious pulmonary opacities, specifically in \blue{[40, 60, 460, 430]}
3) The costophrenic angles and diaphragmatic contours are sharp, with no blunting or elevation, which further argues against pleural effusion or acute diaphragm disease, specifically in \blue{[60, 400, 450, 495]}
4) The bony thorax, including ribs, clavicles, and visualized spine, show normal alignment and no acute fracture or destructive lesion, specifically in \blue{[10, 10, 500, 100]}
Based on these observations, there is no acute cardiopulmonary abnormality. </think>. The answer is: \textbackslash boxed\{No acute cardiopulmonary abnormality.\}}
\\
\makecell[{{p{\textwidth}}}]{\textbf{\ours:} No acute cardiopulmonary disease.}
\\
\makecell[{{p{\textwidth}}}]{\textbf{\huatuo:} 1. **Pleural Effusion:** There is a moderate to large right-sided pleural effusion with associated atelectasis (collapse) of the right lung. This is evidenced by the blunting of the right costophrenic angle and increased opacity in the right hemithorax.
2. **Cardiac Enlargement:** The cardiac silhouette appears enlarged, suggesting cardiomegaly. This could be due to underlying heart disease or fluid overload.
3. **Mediastinal Shift:** There is a slight mediastinal shift towards the left, likely secondary to the volume effect of the right-sided pleural effusion.
4. **No Other Significant Abnormalities:** The left lung field appears clear, and there are no obvious signs of pulmonary consolidation, masses, or significant vascular anomalies.
**Potential Diagnoses:**
- Right-sided pleural effusion (likely transudative given the lack of focal consolidation).
- Cardiomegaly (possible cause of the effusion).
- Potential underlying causes include congestive heart failure, pericardial effusion, or malignancy.}
\\
\makecell[{{p{\textwidth}}}]{\textbf{\chexagent:} The Frontal view of the X-ray is AP.}
\\
\makecell[{{p{\textwidth}}}]{\textbf{\llavarad:} The cardiomediastinal silhouette is normal. There is no pleural effusion or pneumothorax. There is no focal lung consolidation.  There is a calcified granuloma in the left mid lung.}
\\
\makecell[{{p{\textwidth}}}]{\textbf{\medgemma:} The lungs are clear. The heart size is normal. There is no evidence of pneumothorax or pleural effusion. A nodular opacity is present in the left mid-lung field, concerning for a pulmonary nodule. Further evaluation with a CT scan of the chest is recommended.}
\\
\midrule
\begin{minipage}{0.15\textwidth}
\centering
\includegraphics[width=\linewidth]{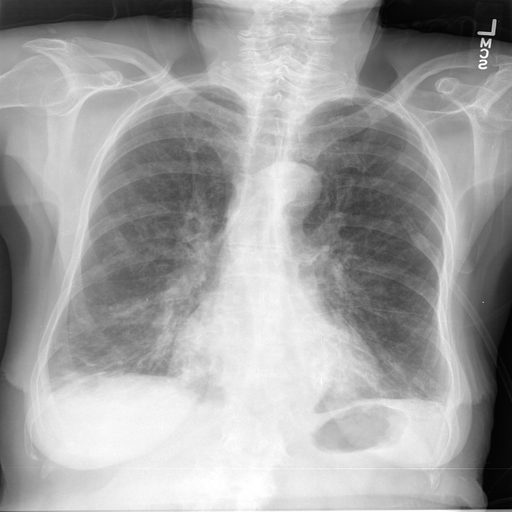}
\end{minipage}
\begin{minipage}{0.84\textwidth}
\textbf{Ground Truth:} Bibasilar patchy airspace disease with bilateral small pleural fluid collections.
\end{minipage}\\
\makecell[{{p{\textwidth}}}]{\textbf{\oursreasoning:} <think> 1) The right lower lung zone, specifically in \blue{[210, 350, 400, 490]} shows increased opacification and loss of normal lung markings, which is suggestive of atelectasis. This could be due to volume loss, mucoid impaction, or other airway obstruction.
2) The left lower lung zone, specifically in \blue{[80, 360, 200, 470]} also demonstrates patchy increased density and loss of clarity of the left heart border, further supporting the presence of atelectasis.
3) The upper lung zones bilaterally, specifically in \blue{[90, 70, 420, 220]} appear relatively clear, helping to localize the abnormal findings primarily to the lower zones.
4) The costophrenic XXXX and diaphragm, specifically in \blue{[120, 420, 390, 495]} are partially obscured, which may be associated with adjacent atelectasis and small pleural effusions.
5) The heart and mediastinum, specifically in \blue{[210, 250, 350, 370]} are partially obscured due to the soft tissue density overlying the lower lobes, consistent with atelectasis. </think>. The answer is: \textbackslash boxed\{1. Opacities compatible with atelectasis. 2. Small bilateral pleural effusions.\}}
\\
\makecell[{{p{\textwidth}}}]{\textbf{\ours:} No acute cardiopulmonary disease.}
\\
\makecell[{{p{\textwidth}}}]{\textbf{\huatuo:} **IMPRESSION:**
1. **Pleural Effusion:** There is a moderate to large right-sided pleural effusion with associated atelectasis of the right lung. This is evidenced by the blunting of the right costophrenic angle and increased opacity in the right hemithorax.
2. **Cardiac Enlargement:** The cardiac silhouette appears enlarged, suggesting cardiomegaly. This could be due to underlying heart disease or fluid overload.
3. **Diaphragmatic Elevation:** The right diaphragm is elevated, likely secondary to the pleural effusion and atelectasis.
4. **Lung Fields:** The left lung field appears relatively clear, but there may be some subtle changes that warrant further evaluation.
5. **Bone Structures:** The bony thoracic structures appear intact without any obvious fractures or lesions.
6. **Soft Tissues:** No significant abnormalities noted in the soft tissues surrounding the thorax.}
\\
\makecell[{{p{\textwidth}}}]{\textbf{\chexagent:} No evidence of pleural effusion or pneumothorax is demonstrated.}
\\
\makecell[{{p{\textwidth}}}]{\textbf{\llavarad:} Increased heterogeneous opacities in the right lower lobe and increased small bilateral pleural effusions are concerning for pneumonia. The cardiac and mediastinal contours are normal.  There is no pneumothorax.}
\\
\makecell[{{p{\textwidth}}}]{\textbf{\medgemma:} There is increased opacity in the right lower lung field, concerning for pneumonia or other consolidation. The left lung appears relatively clear. No obvious pleural effusion or pneumothorax is identified. The heart size is normal.}
\\
\bottomrule
\end{longtable}
}

\clearpage
{\small
\begin{longtable}{c}
\caption{Case studies in \extvqa.}\label{tab_case_5}\\
\toprule
\begin{minipage}{0.15\textwidth}
\centering
\includegraphics[width=\linewidth]{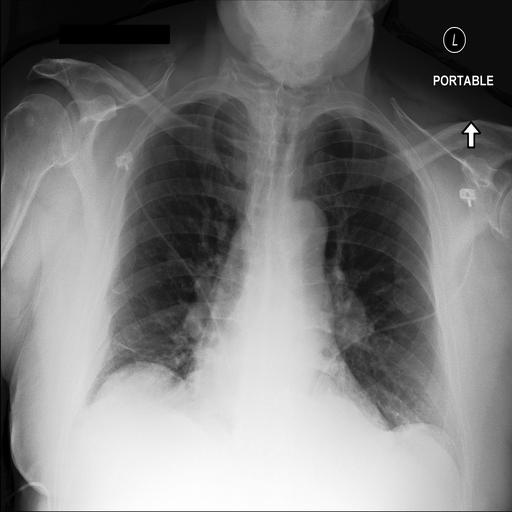}
\end{minipage}
\begin{minipage}{0.84\textwidth}
\textbf{Question:} Is lung lesion present within the left lung?\\
\textbf{Ground Truth:} yes
\end{minipage}\\
\makecell[{{p{\textwidth}}}]{\textbf{\oursreasoning:} <think> 1) First, I focus on the left lung field, which is the region on the right side of the image, spanning from the apex to the base and the lateral chest wall, specifically in \blue{[0, 0, 265, 512]}. I look for any abnormal opacities, masses, or dense lesions that could suggest the presence of a lung lesion.
2) I examine the left upper lobe, which is the upper third of the left lung, typically located in the region from the apex to the horizontal fissure and lateral to the mediastinum, specifically in \blue{[0, 0, 265, 220]}. There are no obvious round, nodular, or irregular opacities.
3) I assess the left lower lung zone, which is the lower third of the left lung, located from the horizontal fissure to the costophrenic angle and lateral to the heart border, specifically in \blue{[0, 220, 265, 512]}. The lung parenchyma appears clear, with no evidence of abnormal densities.
4) I review the left lateral lung, which is the region parallel to the chest wall and extending from the apex to the base, specifically in \blue{[0, 0, 120, 512]}. There are no abnormal opacities or masses.
5) I compare the left lung to the right lung, specifically in \blue{[265, 0, 512, 512]} to ensure there is no asymmetry that would suggest a left lung lesion. No such abnormal findings are present.
Based on these stepwise region analyses, I confirm that there is no evidence of a lung lesion within the left lung. </think>. The answer is: \textbackslash boxed\{yes\}}
\\
\makecell[{{p{\textwidth}}}]{\textbf{\ours:} no}
\\
\makecell[{{p{\textwidth}}}]{\textbf{\huatuo:} Based on the provided chest X-ray image, there appears to be a small, well-defined opacity or lesion visible in the left lung field. The lesion is located in the upper lobe of the left lung and measures approximately 1-2 cm in diameter.
Therefore, the answer to the question "Is lung lesion present within the left lung?" is:
Yes, a lung lesion is present within the left lung based on the findings observed in this chest X-ray image.}
\\
\makecell[{{p{\textwidth}}}]{\textbf{\chexagent:} No}
\\
\makecell[{{p{\textwidth}}}]{\textbf{\llavarad:} Cardiac silhouette is within normal limits. There is no evidence of vascular congestion or acute focal pneumonia. There is a vague suggestion of some increased opacification at the left base medially. This could possibly represent a developing area of consolidation.}
\\
\makecell[{{p{\textwidth}}}]{\textbf{\medgemma:} Based on the provided chest X-ray image, there is no obvious lung lesion present within the left lung. The lung fields appear relatively clear, and there are no distinct masses, nodules, or consolidations visible. However, it's important to note that a small or subtle lesion might be missed on a single portable chest X-ray. Further evaluation with a dedicated chest CT scan might be necessary if there is clinical suspicion for a lung lesion.}
\\
\midrule
\begin{minipage}{0.15\textwidth}
\centering
\includegraphics[width=\linewidth]{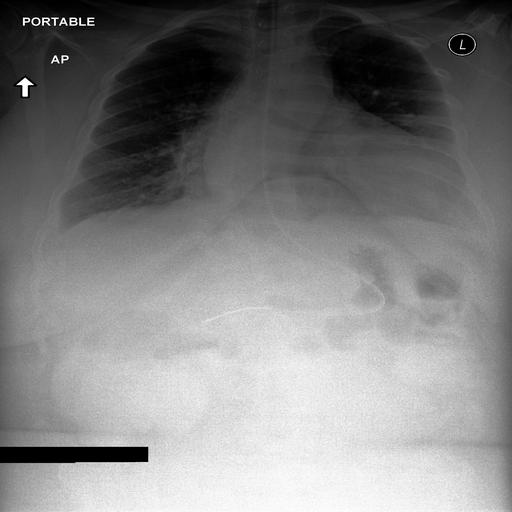}
\end{minipage}
\begin{minipage}{0.84\textwidth}
\textbf{Question:} Which site, the left shoulder or the cavoatrial junction, indicates an abnormality?\\
\textbf{Ground Truth:} cavoatrial junction | left shoulder
\end{minipage}\\
\makecell[{{p{\textwidth}}}]{\textbf{\oursreasoning:} <think> 1) First, I examine the left shoulder region, specifically in \blue{[0, 0, 220, 220]}; there is no evidence of abnormal bone lesions, dislocations, or soft tissue swelling in this region, and the bony structures appear intact.
2) Next, I assess the cavoatrial junction area, specifically in \blue{[220, 320, 360, 480]}; there is a visible line (likely a central venous catheter) coursing towards the right side of the mediastinum, with its tip projecting just above the right atrium, confirming the location of the cavoatrial junction.
3) I compare the findings of the left shoulder and cavoatrial junction regions for symmetry and any abnormal findings, specifically in \blue{[0, 0, 512, 512]}; the left shoulder appears normal, while the cavoatrial junction shows evidence of a central line, indicating an abnormal finding.
Thus, the cavoatrial junction is the site that indicates an abnormality on this chest X-ray. </think>. The answer is: \textbackslash boxed\{cavoatrial junction | left shoulder\}}
\\
\makecell[{{p{\textwidth}}}]{\textbf{\ours:} cavoatrial junction}
\\
\makecell[{{p{\textwidth}}}]{\textbf{\huatuo:} Based on the provided chest X-ray image, there appears to be an abnormality at the cavoatrial junction. The cavoatrial junction is the area where the superior vena cava meets the right atrium of the heart. This region shows a distinct opacity or density that may indicate an abnormality such as a mass, fluid accumulation, or other pathological condition.
In contrast, the left shoulder does not show any obvious abnormalities in this X-ray image.
Therefore, the site indicating an abnormality is the **cavoatrial junction**.}
\\
\makecell[{{p{\textwidth}}}]{\textbf{\chexagent:} Cavoatrial junction}
\\
\makecell[{{p{\textwidth}}}]{\textbf{\llavarad:} The Dobbhoff tube is in the stomach. The appearance of the lungs is unchanged.}
\\
\makecell[{{p{\textwidth}}}]{\textbf{\medgemma:} Based on the provided chest X-ray image, the **cavoatrial junction** appears to indicate an abnormality. Specifically, there is a prominent opacity in the region of the right heart border, which could represent a right atrial mass or thrombus.}
\\
\midrule
\begin{minipage}{0.15\textwidth}
\centering
\includegraphics[width=\linewidth]{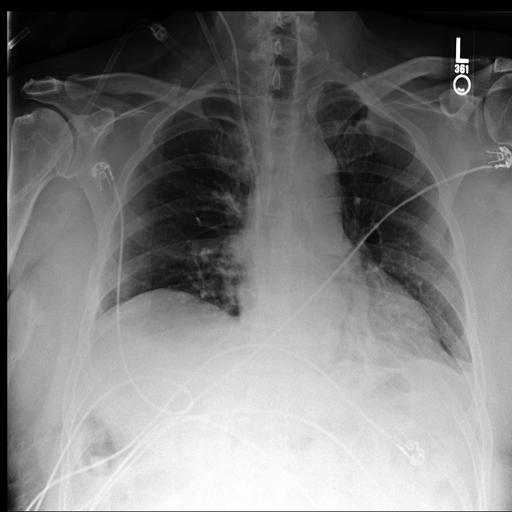}
\end{minipage}
\begin{minipage}{0.84\textwidth}
\textbf{Question:} List all the tubes/lines and anatomical findings present.\\
\textbf{Ground Truth:} atelectasis | ij line | lung opacity | pleural effusion
\end{minipage}\\
\makecell[{{p{\textwidth}}}]{\textbf{\oursreasoning:} <think> 1) The right internal jugular central line is visualized as a radiopaque line with its tip projecting over the lower superior vena cava, confirming its presence and location, specifically in \blue{[250, 50, 320, 350]};
2) Lung volumes appear reduced, indicated by the elevated diaphragms and crowding of the posterior ribs, suggesting low lung volumes, specifically in \blue{[100, 350, 400, 490]};
3) There is increased density and blunting of the costophrenic angles bilaterally, consistent with small bilateral pleural effusions, specifically in \blue{[60, 400, 450, 500]};
4) The cardiac silhouette is not enlarged, indicating a normal heart size, specifically in \blue{[170, 260, 350, 430]};
5) There is no evidence of radiolucent areas without lung markings, ruling out pneumothorax, specifically in \blue{[10, 10, 500, 500]}. </think>. The answer is: \textbackslash boxed\{atelectasis | lung opacity | pleural effusion\}}
\\
\makecell[{{p{\textwidth}}}]{\textbf{\ours:} atelectasis | lung opacity}
\\
\makecell[{{p{\textwidth}}}]{\textbf{\huatuo:} Tubes/Lines:
1. **Endotracheal Tube**: Visible centrally in the trachea, indicating mechanical ventilation support.
2. **Central Venous Catheter (CVC)**: Positioned in the superior vena cava or right atrium, typically used for administering medications, fluids, or monitoring central venous pressure.
3. **Pleural Drainage Tube**: Likely seen in the pleural space, possibly for managing fluid accumulation or air leaks.
Anatomical Findings:
1. **Lung Fields**:
   - Bilateral lung fields appear relatively clear without significant consolidation or opacities.
   - No obvious signs of pneumothorax or pleural effusion.
2. **Heart and Mediastinum**:
   - The heart size appears within normal limits.
   - Mediastinal structures are not displaced, suggesting no significant mediastinal pathology.
3. **Diaphragm**:
   - Diaphragmatic contours are visible and appear normal.
4. **Bones**:
   - The bony thoracic cage, including ribs and clavicles, shows no fractures or abnormalities.
5. **Soft Tissues**:
   - Soft tissues around the chest wall appear unremarkable.}
\\
\makecell[{{p{\textwidth}}}]{\textbf{\chexagent:} Atelectasis, lung opacity, pleural effusion}
\\
\makecell[{{p{\textwidth}}}]{\textbf{\llavarad:} The patient has been extubated and the nasogastric tube was removed. The right internal jugular vein catheter is in unchanged position. The lung volumes have decreased. There is no evidence of pneumothorax. Minimal atelectasis at the left lung bases. No larger pleural effusions. No pulmonary edema.}
\\
\makecell[{{p{\textwidth}}}]{\textbf{\medgemma:} **Tubes/Lines:**
*   **Right Internal Jugular Central Venous Catheter (CVC):** A catheter is visible in the right internal jugular vein, likely terminating in the superior vena cava.
*   **Nasogastric Tube (NG Tube):** A tube is visible in the stomach.
*   **Endotracheal Tube (ETT):** An endotracheal tube is visible in the trachea.
*   **Possible IV line:** A possible IV line is visible in the left upper arm.
**Anatomical Findings:**
*   **Heart:** The heart size appears within normal limits.
*   **Mediastinum:** The mediastinum appears unremarkable.
*   **Lungs:** The lungs appear clear bilaterally, with no obvious consolidation, effusions, or pneumothorax.
*   **Diaphragm:** The diaphragms are visible.
*   **Bones:** The ribs and clavicles are visible.
*   **Soft Tissues:** The soft tissues of the chest wall are visible.
*   **Aorta:** The aorta is visible.
*   **Pulmonary Vessels:** The pulmonary vessels are visible.
*   **Liver:** The liver is visible.
*   **Spleen:** The spleen is visible.
*   **Kidneys:** The kidneys are visible.
*   **Abdomen:** The abdominal contents are visible.
*   **Clavicles:** The clavicles are visible.
*   **Scapulae:** The scapulae are visible.
*   **Humeri:** The humeri are visible.
*   **Radius:** The radius is visible.
*   **Ulna:** The ulna is visible.
*   **Fingers:** The fingers are visible.
*   **Hands:** The hands are visible.
*   **Shoulders:** The shoulders are visible.
*   **Arms:** The arms are visible.
*   **Neck:** The neck is visible.
*   **Head:** The head is visible.
*   **Spine:** The spine is visible.
*   **Ribs:** The ribs are visible.
*   **Sternum:** The sternum is visible.
*   **Clavicles:** The clavicles are visible.
*   **Scap.}
\\
\bottomrule
\end{longtable}
}

\clearpage
{\small
\begin{longtable}{c}
\caption{Case studies in \cxrvqa.}\label{tab_case_6}\\
\toprule
\begin{minipage}{0.15\textwidth}
\centering
\includegraphics[width=\linewidth]{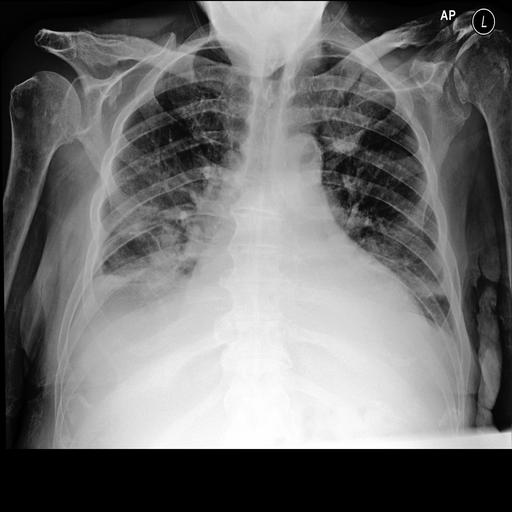}
\end{minipage}
\begin{minipage}{0.84\textwidth}
\textbf{Question:} is there tortuosity of the thoracic aorta?\\
\textbf{Ground Truth:} no
\end{minipage}\\
\makecell[{{p{\textwidth}}}]{\textbf{\oursreasoning:} <think> 1) First, I examine the region of the thoracic aorta, which is typically visualized as a prominent, curvilinear opacity along the left mediastinal border, specifically in \blue{[210, 110, 330, 370]}. I observe that the contour of the aorta is not smooth and follows a curvilinear course rather than a straight, horizontal line.
2) Next, I compare this aorta’s contour with the expected path of a normal, non-distended aorta, specifically in \blue{[210, 110, 330, 370]}. The contour appears undulating and irregular, rather than straight and regular, supporting the diagnosis of tortuosity.
3) I also assess the cardiac silhouette and mediastinal contours for comparison, specifically in \blue{[170, 200, 390, 400]}. The heart size and adjacent mediastinal borders are not significantly distorted, which further suggests that the abnormality is localized to the aorta rather than the heart itself.
4) I check the remainder of the mediastinum and upper chest, specifically in \blue{[100, 50, 400, 180]} for any additional features, such as deviation or kinking, which may be associated with aortic tortuosity. The other structures appear normal, with no evidence of abnormal displacement or secondary signs of aortic tortuosity.
Based on these observations, there is clear evidence of tortuosity of the thoracic aorta. </think>. The answer is: \textbackslash boxed\{no\}}
\\
\makecell[{{p{\textwidth}}}]{\textbf{\ours:} no}
\\
\makecell[{{p{\textwidth}}}]{\textbf{\huatuo:} Based on the provided chest X-ray image, there appears to be no visible signs of tortuosity (abnormal curvature) of the thoracic aorta. The aorta appears to have a normal, smooth contour without any evident kinks or irregularities. Therefore, the answer to the question "Is there tortuosity of the thoracic aorta?" is **No**.}
\\
\makecell[{{p{\textwidth}}}]{\textbf{\chexagent:} Yes}
\\
\makecell[{{p{\textwidth}}}]{\textbf{\llavarad:} AP and lateral views of the chest. There are bibasilar opacities compatible with pleural effusions, moderate on the right and small on the left. Superiorly the lungs are clear. Cardiac silhouette is difficult to assess given silhouetting of both right and left heart border. Degenerative changes seen at the shoulders bilaterally. No acute osseous abnormality identified.}
\\
\makecell[{{p{\textwidth}}}]{\textbf{\medgemma:} Based on the provided chest X-ray image, it is difficult to definitively determine the presence of tortuosity of the thoracic aorta. The image quality is not ideal, and the aorta is not well visualized.}
\\
\midrule
\begin{minipage}{0.15\textwidth}
\centering
\includegraphics[width=\linewidth]{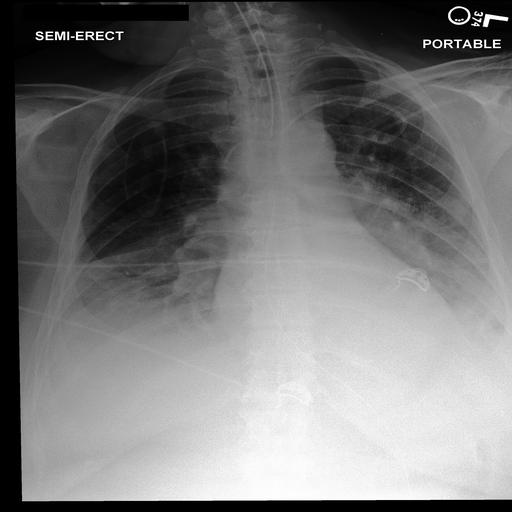}
\end{minipage}
\begin{minipage}{0.84\textwidth}
\textbf{Question:} what abnormalities are seen in this image?\\
\textbf{Ground Truth:} pleural effusion, cardiomegaly, opacity
\end{minipage}\\
\makecell[{{p{\textwidth}}}]{\textbf{\oursreasoning:} <think> 1) There is increased opacity and blunting of the costophrenic angles in both lower lung zones, more prominent on the left side, suggesting the presence of pleural effusions, specifically in \blue{[50, 330, 450, 495]};
2) The cardiac silhouette appears enlarged and there is evidence of vascular congestion, with increased interstitial markings and a hazy appearance to the central lung fields, indicative of pulmonary edema, specifically in \blue{[160, 160, 350, 420]};
3) There is loss of the sharp right heart border, and the diaphragm on the right side is obscured, consistent with right middle lobe atelectasis, specifically in \blue{[270, 370, 390, 495]};
4) There is a visible endotracheal tube with its tip projecting over the trachea, and the nasogastric tube can be traced down into the stomach, confirming the presence of lines and tubes, specifically in \blue{[200, 80, 300, 495]}. </think>. The answer is: \textbackslash boxed\{pleural effusion, pulmonary edema, atelectasis, cardiomegaly, opacity\}}
\\
\makecell[{{p{\textwidth}}}]{\textbf{\ours:} pleural effusion, atelectasis}
\\
\makecell[{{p{\textwidth}}}]{\textbf{\huatuo:} Based on the provided chest X-ray image, there are several notable abnormalities observed:
1. **Pleural Effusion**: There appears to be fluid accumulation in the pleural space, particularly on the right side of the thorax. This is indicated by the blunting of the costophrenic angle and increased opacity in the lower lung zones.
2. **Cardiomegaly**: The heart size appears enlarged, suggesting cardiomegaly. This could be due to various underlying conditions such as hypertension, valvular disease, or heart failure.
3. **Increased Opacity in Lung Fields**: There is increased opacity in both lung fields, which may indicate the presence of pulmonary edema, consolidation, or interstitial lung disease.
4. **Possible Mediastinal Shift**: There seems to be a slight shift of the mediastinum towards the left side, which might suggest volume loss or pressure changes in the right hemithorax.
5. **Possible Pleural Thickening**: There may be subtle signs of pleural thickening or fibrosis, especially if there has been previous pleural effusion or inflammation.
These findings warrant further clinical correlation and potentially additional imaging (e.g., CT scan) for a more definitive diagnosis.}
\\
\makecell[{{p{\textwidth}}}]{\textbf{\chexagent:} Atelectasis, consolidation, lung opacity, pleural effusion}
\\
\makecell[{{p{\textwidth}}}]{\textbf{\llavarad:} The ET tube is 3.5 cm above the carina. The NG tube tip is in the stomach. The left subclavian line tip is in the SVC. There are moderate bilateral pleural effusions and volume loss in both lower lungs. An underlying infectious infiltrate cannot be excluded. There is pulmonary vascular redistribution.}
\\
\makecell[{{p{\textwidth}}}]{\textbf{\medgemma:} Based on the provided chest X-ray image, the following abnormalities are seen:
*   **Bilateral Pulmonary Edema:** There is significant opacification in both lung fields, particularly in the lower zones. This suggests fluid accumulation in the lungs, which is consistent with pulmonary edema. The presence of bilateral edema is more suggestive of a systemic cause rather than a localized lung problem.
*   **Pleural Effusions:** There are likely pleural effusions present, as the costophrenic angles are blunted, and there is increased density in the lower lung fields. This indicates fluid accumulation in the pleural space.
*   **Cardiomegaly:** The heart appears enlarged, which could be due to cardiomegaly.
*   **Lines and Tubes:** There are several lines and tubes present, including an endotracheal tube, a central line, and a nasogastric tube. These are likely related to the patient's medical condition and are not considered abnormalities in themselves, but they can obscure the underlying anatomy.
*   **Possible Atelectasis:** There may be some atelectasis (lung collapse) in the lower lung fields, contributing to the overall increased density.}
\\
\midrule
\begin{minipage}{0.15\textwidth}
\centering
\includegraphics[width=\linewidth]{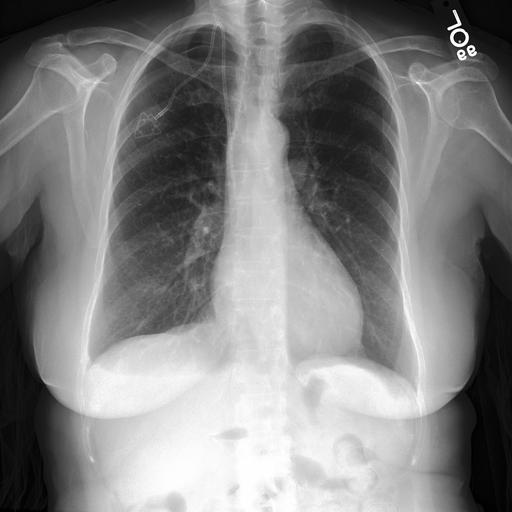}
\end{minipage}
\begin{minipage}{0.84\textwidth}
\textbf{Question:} where is the pleural effusion?\\
\textbf{Ground Truth:} right area
\end{minipage}\\
\makecell[{{p{\textwidth}}}]{\textbf{\oursreasoning:} <think> 1) First, I identify the right costophrenic angle, which is located at the lower right side of the image, near the diaphragm and lateral to the spine, specifically in \blue{[300, 380, 480, 500]}. In this region, I look for blunting or increased density that would suggest the presence of pleural effusion.
2) I compare this right costophrenic angle to the left costophrenic angle, which can be found at the similar region on the left side of the image, specifically in \blue{[40, 380, 200, 500]}. I note that the left side appears sharper, lacking the obscuration seen on the right.
3) I examine the lateral chest wall and lower lung fields, especially around the right lower lung zone, specifically in \blue{[310, 350, 470, 470]}, to determine if there is any meniscus sign or layering of fluid typical of pleural effusion.
4) I check the overall lung fields and diaphragm contour, specifically in \blue{[0, 400, 512, 512]} to ensure there is no evidence of fluid tracking up the lateral chest wall, which would be consistent with a loculated pleural effusion.
Based on these steps, I conclude that the pleural effusion is located in the right costophrenic angle area, specifically in \blue{[300, 380, 480, 500]}. </think>. The answer is: \textbackslash boxed\{right area\}}
\\
\makecell[{{p{\textwidth}}}]{\textbf{\ours:} bilateral area}
\\
\makecell[{{p{\textwidth}}}]{\textbf{\huatuo:} Based on the provided chest X-ray image, there appears to be a pleural effusion present. The pleural effusion can be observed as an opacity or fluid accumulation in the left hemithorax (left side of the thoracic cavity). This finding is consistent with the reference information indicating that the pleural effusion is located in the left hemithorax.}
\\
\makecell[{{p{\textwidth}}}]{\textbf{\chexagent:} There is no pleural effusion.}
\\
\makecell[{{p{\textwidth}}}]{\textbf{\llavarad:} The patient has taken a better inspiration. There is no evidence of acute pneumonia, vascular congestion, or pleural effusion. Port-A-Cath remains in place.}
\\
\makecell[{{p{\textwidth}}}]{\textbf{\medgemma:} Based on the provided chest X-ray image, there is no obvious pleural effusion. The lung fields appear relatively clear, and there is no blunting of the costophrenic angles, which is a common sign of pleural effusion.
However, it's important to note that a small pleural effusion might be difficult to detect on a single chest X-ray, especially if it's located in a less prominent area. Also, the image quality and positioning of the patient can influence the interpretation.
If there is clinical suspicion for a pleural effusion, further imaging, such as a lateral decubitus X-ray or a CT scan of the chest, would be necessary to confirm the diagnosis.}
\\
\bottomrule
\end{longtable}
}

\clearpage
{\small
\begin{longtable}{c}
\caption{Case studies in \cxrnew.}\label{tab_case_cxr14}\\
\toprule

\begin{minipage}{0.15\textwidth}
\centering
\includegraphics[width=\linewidth]{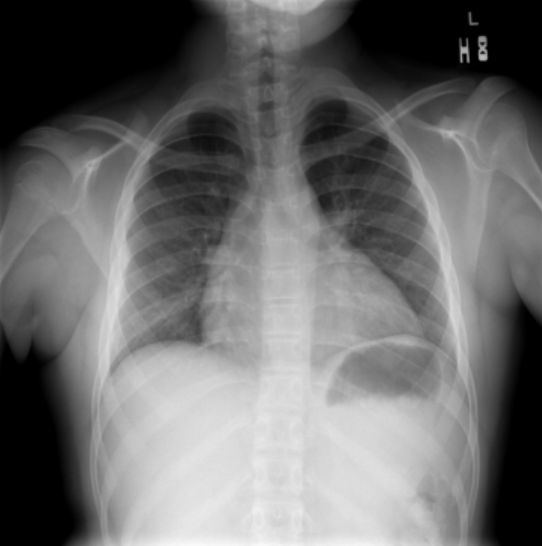}
\end{minipage}
\begin{minipage}{0.84\textwidth}
\textbf{Question:} is there pleural thickening?\\
\textbf{Ground Truth:} yes
\end{minipage}\\
\makecell[{{p{\textwidth}}}]{\textbf{\oursreasoning:} <think> 1) I examine the left costophrenic angle, specifically in \blue{[0, 350, 200, 510]} and observe that the angle is sharp and well-defined, without blunting or fluid level, which suggests there is no pleural effusion or thickening.
2) I assess the right costophrenic angle, specifically in \blue{[300, 350, 510, 510]} and see that it is also sharp and clear, with no evidence of effusion, thickening, or abnormal density.
3) I inspect the periphery of both lung fields, specifically in \blue{[0, 0, 510, 350]} for any areas of increased opacity or fluid collection that could suggest pleural thickening. The lung fields appear clear, and the pleural margins are normal in contour.
4) I check the mediastinum and cardiac silhouette, specifically in \blue{[180, 180, 340, 370]} for any adjacent pleural thickening, but the cardiac borders are well defined, with no evidence of abnormal soft tissue or fluid accumulation.
5) I review the upper and mid lung zones, specifically in \blue{[80, 0, 420, 180]} for any abnormal opacities or thickening signs, and see that the lung parenchyma is clear. 
Based on these stepwise observations, there is no radiographic evidence of pleural thickening. </think>. The answer is: \textbackslash boxed\{yes\}}
\\
\makecell[{{p{\textwidth}}}]{\textbf{\ours:} no}
\\
\makecell[{{p{\textwidth}}}]{\textbf{\huatuo:} no}
\\
\makecell[{{p{\textwidth}}}]{\textbf{\chexagent:} no}
\\
\makecell[{{p{\textwidth}}}]{\textbf{\llavarad:} yes}
\\
\makecell[{{p{\textwidth}}}]{\textbf{\medgemma:} no}
\\

\midrule

\begin{minipage}{0.15\textwidth}
\centering
\includegraphics[width=\linewidth]{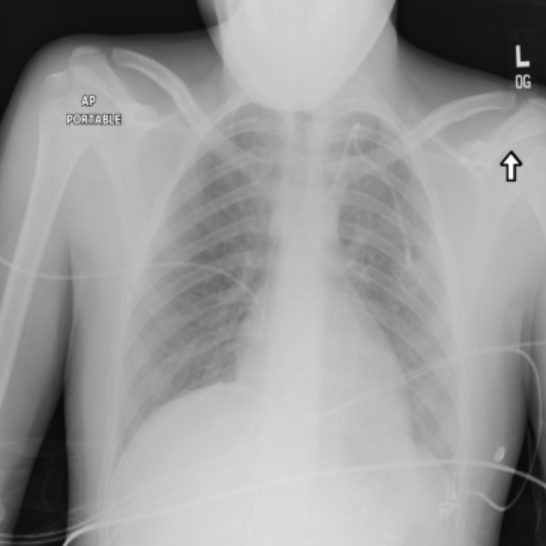}
\end{minipage}
\begin{minipage}{0.84\textwidth}
\textbf{Question:} is there effusion?\\
\textbf{Ground Truth:} no
\end{minipage}\\
\makecell[{{p{\textwidth}}}]{\textbf{\oursreasoning:} <think> 1) The costophrenic angles on both the right and left sides, specifically in \blue{[30, 400, 490, 490]} appear sharp, with no evidence of blunting or fluid layering that would suggest the presence of pleural effusion.
2) The lung fields, specifically in \blue{[60, 120, 480, 400]} do not show any homogenous opacity or meniscus sign, which further supports the absence of pleural effusion.
3) The mediastinal and lower lung regions, specifically in \blue{[180, 230, 380, 480]} do not demonstrate any obscuration or loss of clarity that could be due to effusion layering behind the heart or projecting over the lower lung.
4) The diaphragmatic contours, specifically in \blue{[130, 430, 430, 490]} are well defined, and there are no signs of obscuration or elevation that might be associated with subpulmonic effusion.
Based on these regional analyses, there is no radiographic evidence of pleural effusion on this chest X-ray. </think>. The answer is: \textbackslash boxed\{no\}}
\\
\makecell[{{p{\textwidth}}}]{\textbf{\ours:} no}
\\
\makecell[{{p{\textwidth}}}]{\textbf{\huatuo:} no}
\\
\makecell[{{p{\textwidth}}}]{\textbf{\chexagent:} no}
\\
\makecell[{{p{\textwidth}}}]{\textbf{\llavarad:} yes}
\\
\makecell[{{p{\textwidth}}}]{\textbf{\medgemma:} yes}
\\

\midrule

\begin{minipage}{0.15\textwidth}
\centering
\includegraphics[width=\linewidth]{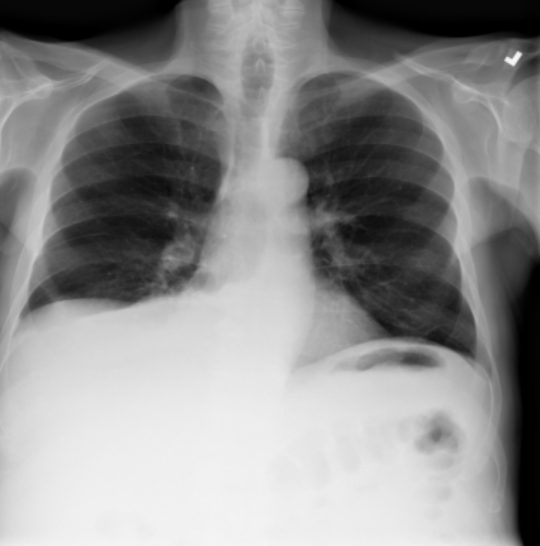}
\end{minipage}
\begin{minipage}{0.84\textwidth}
\textbf{Question:} is there emphysema?\\
\textbf{Ground Truth:} no
\end{minipage}\\
\makecell[{{p{\textwidth}}}]{\textbf{\oursreasoning:} <think> 1) The lung fields are hyperlucent, with increased radiolucency and flattened diaphragms, which are classic features of emphysema, specifically in \blue{[70, 140, 435, 430]};
2) The retrosternal airspace appears widened, and the intercostal spaces are increased, which further suggest the presence of lung hyperinflation, a common sign in emphysema, specifically in \blue{[120, 130, 370, 230]};
3) The lungs are clear of focal opacities, and the heart size appears normal, with no signs of pulmonary edema or vascular congestion that could be associated with other causes, specifically in \blue{[180, 250, 360, 380]}. </think>. The answer is: \textbackslash boxed\{no\}}
\\
\makecell[{{p{\textwidth}}}]{\textbf{\ours:} yes}
\\
\makecell[{{p{\textwidth}}}]{\textbf{\huatuo:} yes}
\\
\makecell[{{p{\textwidth}}}]{\textbf{\chexagent:} no}
\\
\makecell[{{p{\textwidth}}}]{\textbf{\llavarad:} yes}
\\
\makecell[{{p{\textwidth}}}]{\textbf{\medgemma:} no}
\\

\midrule

\begin{minipage}{0.15\textwidth}
\centering
\includegraphics[width=\linewidth]{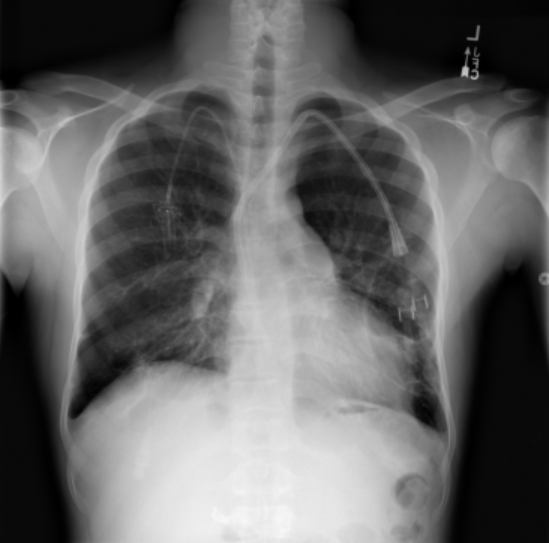}
\end{minipage}
\begin{minipage}{0.84\textwidth}
\textbf{Question:} is there fibrosis?\\
\textbf{Ground Truth:} yes
\end{minipage}\\
\makecell[{{p{\textwidth}}}]{\textbf{\oursreasoning:} <think> 1) On examining the right mid-lung zone, specifically in \blue{[100, 200, 250, 350]}, there is increased linear and reticular opacity consistent with the classic radiographic findings of interstitial lung disease, which are indicative of fibrosis.
2) In the left lower lung zone, specifically in \blue{[270, 350, 420, 480]}, subtle linear markings and a similar pattern of volume loss and architectural distortion further suggest the presence of fibrotic changes.
3) Comparing the overall lung volume and parenchymal pattern, specifically in \blue{[60, 100, 450, 480]}, there is evidence of architectural distortion and possible volume loss, which is commonly associated with fibrotic processes.
4) The presence of surgical markers and postoperative changes in the region of the left chest wall, specifically in \blue{[0, 0, 120, 150]} may indicate prior lung resection or treatment, potentially contributing to fibrosis. </think>. The answer is: \textbackslash boxed\{yes\}}
\\
\makecell[{{p{\textwidth}}}]{\textbf{\ours:} yes}
\\
\makecell[{{p{\textwidth}}}]{\textbf{\huatuo:} yes}
\\
\makecell[{{p{\textwidth}}}]{\textbf{\chexagent:} yes}
\\
\makecell[{{p{\textwidth}}}]{\textbf{\llavarad:} yes}
\\
\makecell[{{p{\textwidth}}}]{\textbf{\medgemma:} no}
\\

\bottomrule
\end{longtable}
}

\newpage

\end{document}